\documentclass[]{fairmeta}
\usepackage[disable]{todonotes}
\usepackage{enumitem}
\usepackage{tabularx} 
\usepackage{listings}
\usepackage{multirow}
\usepackage{makecell}
\usepackage{graphicx}
\usepackage{wrapfig}
\usepackage{tabularx}
\usepackage{textcomp}
\usepackage{stfloats}
\usepackage{url}
\usepackage{verbatim}
\usepackage{titlesec}
\usepackage{tocloft}
\usepackage{adjustbox}
\usepackage{multirow}
\usepackage{pifont}
\usepackage{tikz}
\usepackage{comment}
\usepackage{amsmath,amssymb} 
\usepackage{colortbl}  
\usepackage{color}
\usepackage{booktabs} 
\usepackage{hyperref}
\usepackage{graphicx}    
\usepackage{subcaption} 
\usepackage{multirow} 
\usepackage{booktabs} 
\usepackage{subcaption} 
\RequirePackage{xspace}
\makeatletter
\DeclareRobustCommand\onedot{\futurelet\@let@token\@onedot}
\def\@onedot{\ifx\@let@token.\else.\null\fi\xspace}
\usepackage[most]{tcolorbox}
\usepackage{xcolor}
\usepackage{booktabs}
\usepackage{multirow}
\usepackage{makecell}
\usepackage{diagbox}
\usepackage{adjustbox}
\usepackage[table]{xcolor}
\usepackage{xltabular}
\setcitestyle{square,comma,numbers,sort&compress}

\makeatother

\definecolor{adptorange}{RGB}{248, 205, 172}
\definecolor{cmpblue}{RGB}{189, 215, 238}
\definecolor{cmpblue}{RGB}{189, 215, 238}
\definecolor{rynn}{RGB}{108,92,186} 

\definecolor{our_red}{RGB}{232,157,160}
\definecolor{our_blue}{RGB}{136,206,230}
\definecolor{our_orange}{RGB}{246,200,168}
\definecolor{our_green}{RGB}{178,211,164}

\definecolor{attn_code0}{RGB}{247,215,200}
\definecolor{attn_code1}{RGB}{238,169,139}
\definecolor{mlp_code0}{RGB}{204,201,221}
\definecolor{mlp_code1}{RGB}{102,95,153}

\definecolor{token_blue}{RGB}{84, 120, 140}

\definecolor{myblue}{RGB}{233, 241, 249}
\definecolor{mygray}{RGB}{99, 110, 114}
\definecolor{myred}{RGB}{255, 118, 117}
\definecolor{myyellow}{RGB}{255, 234, 167}
\definecolor{mygreen}{RGB}{216, 226, 204}
\definecolor{mypurple}{RGB}{162, 155, 254}
\definecolor{mybrown}{RGB}{215, 190, 154}
\definecolor{myorange}{RGB}{255, 220, 190}

\usepackage{tabularx} 
\usepackage{pifont}       
\usepackage{bbding}       
\usepackage{fontawesome}
\usepackage{xspace}

\usepackage{float}

\newlength\savewidth

\newcolumntype{x}[1]{>{\centering\arraybackslash}p{#1pt}}
\newcolumntype{y}[1]{>{\raggedright\arraybackslash}p{#1pt}}
\newcolumntype{z}[1]{>{\raggedleft\arraybackslash}p{#1pt}}

\renewcommand{\paragraph}[1]{\vspace{1mm}\noindent\textbf{#1}}
\usepackage{colortbl}
\usepackage{xcolor}
\usepackage{wrapfig}
\usepackage{amssymb}

\renewcommand{\paragraph}[1]{\vspace{1.25mm}\noindent\textbf{#1}}

\usepackage{algorithm}
\usepackage{listings}

\definecolor{codeblue}{rgb}{0.25, 0.5, 0.5}
\definecolor{codekw}{rgb}{0.35, 0.35, 0.75}
\lstdefinestyle{Pytorch}{
    language = Python,
    backgroundcolor = \color{white},
    basicstyle = \fontsize{9pt}{8pt}\selectfont\ttfamily\bfseries,
    columns = fullflexible,
    aboveskip=1pt,
    belowskip=1pt,
    breaklines = true,
    captionpos = b,
    commentstyle = \color{codeblue},
    keywordstyle = \color{codekw},
}

\definecolor{green}{HTML}{009000}
\definecolor{red}{HTML}{ea4335}

\newcommand{\huggingface}{\raisebox{-1.5pt}{\includegraphics[height=1.05em]{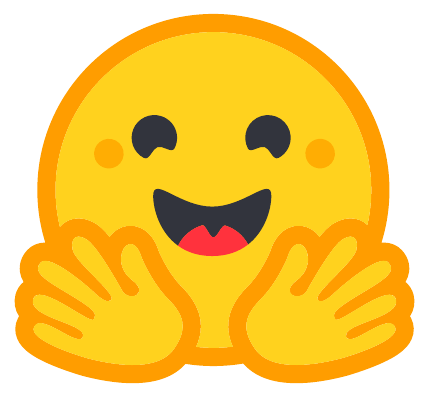}}\xspace}
\newcommand{\github}{\raisebox{-1.5pt}{\includegraphics[height=1.05em]{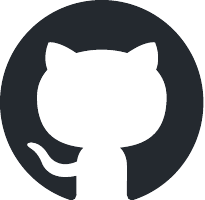}}\xspace}
\newcommand{\homepage}{\raisebox{-1.5pt}{\includegraphics[height=1.05em]{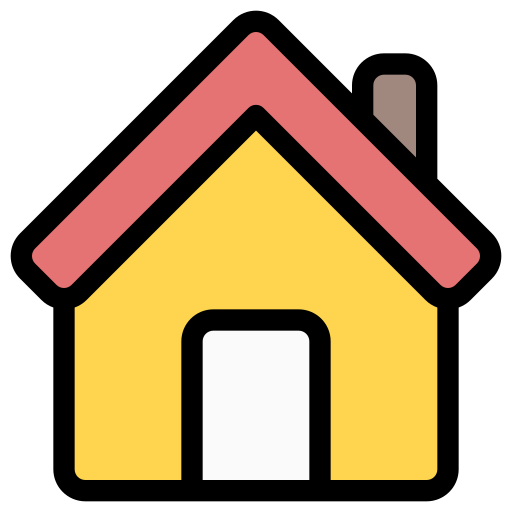}}\xspace}
\newcommand{\modelscope}{\raisebox{-1pt}{\includegraphics[height=0.95em]{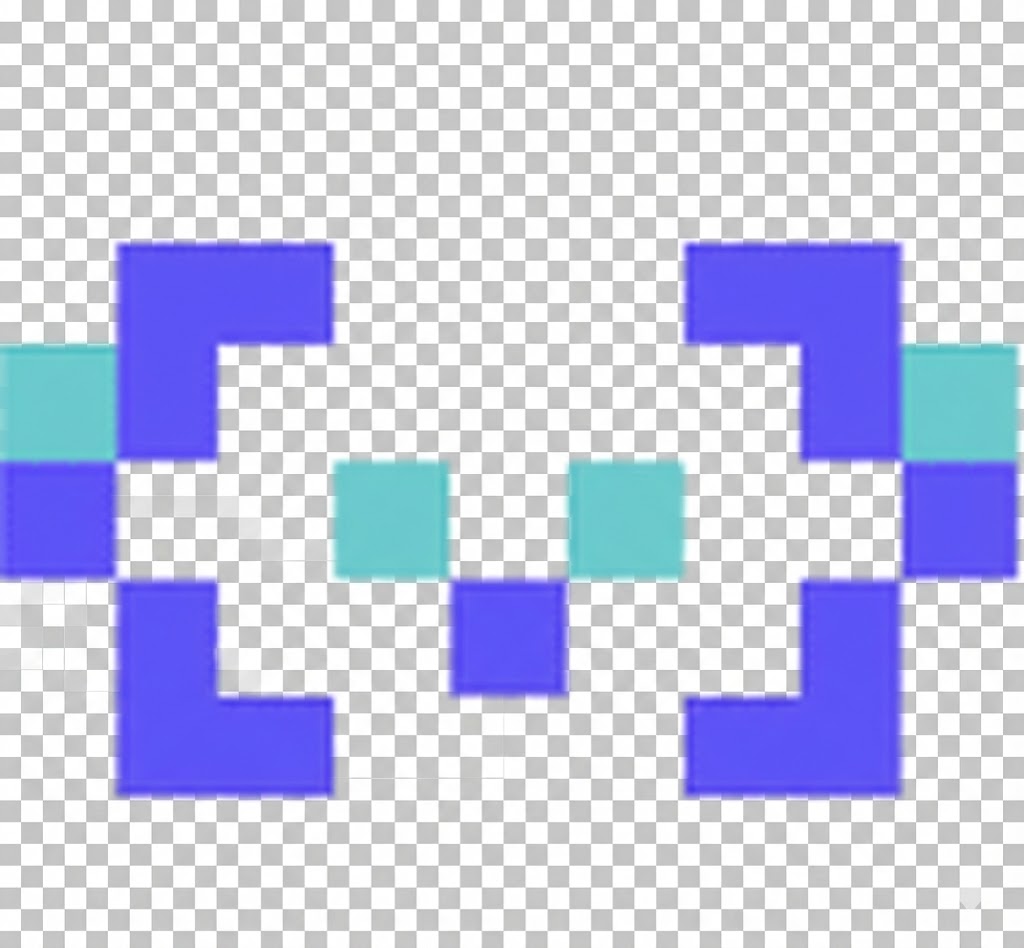}}\xspace}

\title{RynnBrain: Open Embodied Foundation Models}

\author[*,\dagger,\ddagger]{Ronghao Dang}
\author[*,\dagger]{Jiayan Guo}
\author[*]{Bohan Hou}
\author[*]{Sicong Leng}
\author[*,\dagger]{Kehan Li}
\author[*,\dagger]{Xin Li}
\author[*]{Jiangpin Liu}
\author[*]{Yunxuan Mao}
\author[*]{Zhikai Wang}
\author[*]{Yuqian Yuan}
\author[*]{Minghao Zhu}
\author{Xiao Lin}
\author{Yang Bai}
\author{Qian Jiang}
\author{Yaxi Zhao}
\author{Minghua Zeng}
\author{Junlong Gao}
\author{Yuming Jiang}
\author{Jun Cen}
\author{Siteng Huang}
\author{Liuyi Wang}
\author{Wenqiao Zhang}
\author{Chengju Liu}
\author{Jianfei Yang}
\author{Shijian Lu}
\author{Deli Zhao}

\affiliation{DAMO Academy, Alibaba Group\\}
\contribution[*]{Core contributors in alphabetical order}
\contribution[\dagger]{Project lead}
\contribution[\ddagger]{Correspondence}

\abstract{
Despite rapid progress in multimodal foundation models, embodied intelligence community still lacks a unified, physically grounded foundation model that integrates perception, reasoning, and planning within real-world spatial-temporal dynamics.
We introduce \textbf{RynnBrain}, an open-source spatiotemporal foundation model for embodied intelligence. RynnBrain strengthens four core capabilities in a unified framework: comprehensive egocentric understanding, diverse spatiotemporal localization, physically grounded reasoning, and physics-aware planning.
The RynnBrain family comprises three foundation model scales (2B, 8B, and 30B-A3B MoE) and four post-trained variants tailored for downstream embodied tasks (i.e., \textbf{RynnBrain-Nav}, \textbf{RynnBrain-Plan}, and \textbf{RynnBrain-VLA}) or complex spatial reasoning tasks (i.e., \textbf{RynnBrain-CoP}). 
In terms of extensive evaluations on 20 embodied benchmarks and 8 general vision understanding benchmarks, our RynnBrain foundation models largely outperform existing embodied foundation models by a significant margin.
The post-trained model suite further substantiates two key potentials of the RynnBrain foundation model: (i) enabling physically grounded reasoning and planning, and (ii) serving as a strong pretrained backbone that can be efficiently adapted to diverse embodied tasks.

\begin{center}
    \renewcommand{\arraystretch}{1.2}
    \begin{tabular}{ll}
        \homepage  & \url{https://alibaba-damo-academy.github.io/RynnBrain.github.io} \\
        \github  & \url{https://github.com/alibaba-damo-academy/RynnBrain}\\
        \huggingface & \url{https://huggingface.co/collections/Alibaba-DAMO-Academy/rynnbrain} \\
        \modelscope  & \url{https://www.modelscope.cn/collections/DAMO_Academy/RynnBrain}\\
    \end{tabular}
\end{center}

}

\date{\today}

\begin{document}
\thispagestyle{firstheader}
\maketitle
\pagestyle{empty}
\setcounter{tocdepth}{3} 


\section{Introduction} \label{sec:introduction}
\noindent 



The advent of advanced robotic embodiments~\cite{survey_humanoid, SoFTA} and general-purpose vision-language models (VLMs)~\cite{gpt4o,gemini} has created a growing anticipation for versatile robots capable of adaptively performing diverse and complex tasks, which is often referred to as ``embodied intelligence''. A central challenge in embodied intelligence is achieving behavioral and cognitive generalization: enabling robotic agents to transfer knowledge across environments, tasks, and interaction regimes.
 
Despite the strong generalization capabilities, existing VLMs are not intrinsically grounded in physical dynamics and thus struggle with spatio-temporal consistency, physical reasoning, and actionable planning. Conversely, embodied models trained primarily on action-centric data often sacrifice high-level semantic abstraction and lose the broad generalization capabilities inherited from large-scale multimodal pretraining. We argue that progress toward general-purpose embodied intelligence requires a unified foundation model that preserves the semantic breadth of VLMs while being explicitly structured around physical space, temporal dynamics, and embodiment constraints. Such an embodied foundation model should serve as a high-level cognitive ``brain'' for perception, reasoning, and decision-making, while remaining adaptable to downstream control systems. This report primarily examines how to develop a generalizable foundation model for embodied tasks and explores its generalization capacity and post-training potential in multiple dimensions.

\begin{figure}[!htbp]
\centering
\includegraphics[width=\textwidth]{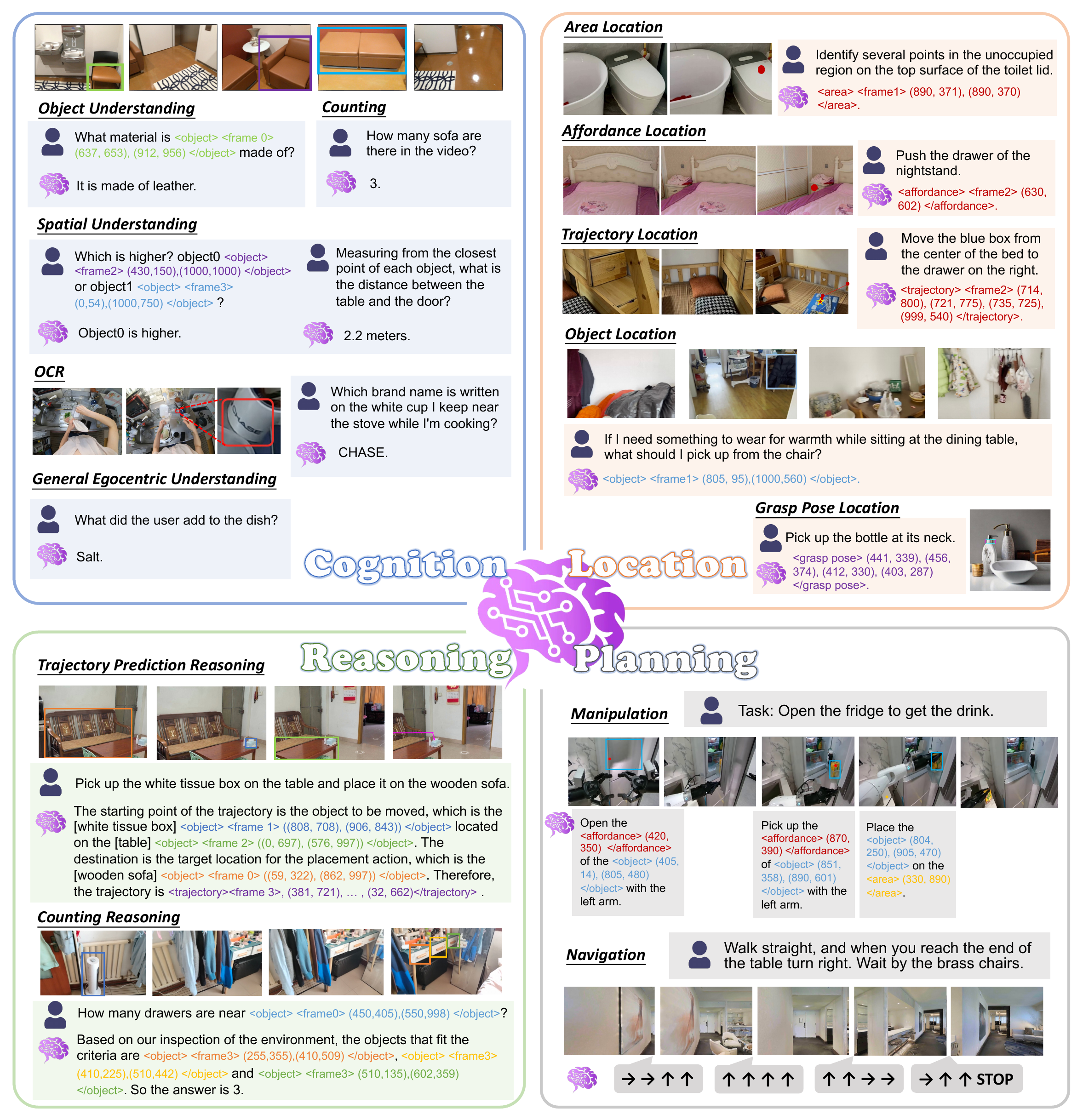}
\caption{Overview of the RynnBrain embodied foundation model. RynnBrain integrates four core capabilities: egocentric cognition, spatio-temporal localization, physically grounded reasoning, and physics-aware planning.
On the input side, RynnBrain processes multimodal signals including images, videos, and spatio-temporal coordinates. On the output side, it jointly produces natural language and explicit spatial grounding primitives such as points, bounding boxes, and trajectories, enabling coherent perception, reasoning, and planning in physical environments.}
\label{fig:intro}
\end{figure}




Several recent efforts~\cite{robobrain2.0, Robix, pelican-VL, MiMo-Embodied, azzolini2025cosmos} have initiated exploration of embodied foundation models. For instance, RoboBrain 2.0~\cite{robobrain2.0} unifies understanding, localization, and planning within a single VLM to facilitate complex embodied tasks, while Robix~\cite{Robix} emphasizes more natural human–robot interaction during execution.
Despite these advances, existing embodied ``brain'' models exhibit three key limitations. First, their egocentric cognitive capabilities remain narrow, as training is typically confined to limited task categories or perception modalities, restricting robustness in complex environments. Second, spatial reasoning is often grounded in static image inputs, lacking coherent spatio-temporal representations necessary for global scene awareness and mobile manipulation. Third, high-level reasoning and planning are frequently conducted in a purely textual space, leading to hallucinations and inconsistencies with physical constraints.



To advance the role of embodied ``brains'' in complex real-world tasks, we propose \textbf{RynnBrain}, a spatio-temporal foundation model explicitly grounded in physical environments. As illustrated in \Cref{fig:intro}, RynnBrain demonstrates robust capabilities in four key dimensions:

\begin{enumerate}
    \item \textbf{Comprehensive egocentric understanding}: RynnBrain excels in spatial comprehension, embodied question answering, egocentric counting, egocentric OCR, etc. Notably, it also introduces fine-grained video understanding—a capability previously overlooked by existing embodied brains.

    \item \textbf{Diverse spatio-temporal localization}: RynnBrain can locate objects, target areas, and even predict trajectories across its entire episodic memory, thereby endowing robots with global spatial awareness.

    \item \textbf{Physically grounded reasoning}: Instead of conventional textual reasoning, RynnBrain employs an interleaved reasoning strategy that alternates between textual and spatial localization, ensuring that its reasoning traces are firmly grounded in the physical environment.

    \item \textbf{Physics-aware planning}: To provide downstream policy models with more accurate planning instructions, RynnBrain integrates the location information of affordance, areas, and objects directly into its planning outputs. Consequently, even highly intricate and fine-grained tasks can be effectively addressed within our hierarchical system architecture.
\end{enumerate}



We build RynnBrain on top of Qwen3-VL~\cite{qwen3vl}. To accommodate varying computational resource constraints, we release two dense variants (2B and 8B) and one mixture-of-experts (MoE) model (30B-A3B). Given comparable inference latency, RynnBrain surpasses all existing embodied brain models in terms of comprehension, localization, and planning capabilities.
Beyond foundational pretraining, we explore four post-training directions: RynnBrain‑CoP, RynnBrain‑Nav, RynnBrain‑Plan, and RynnBrain-VLA. RynnBrain-CoP introduces chain-of-point reasoning, an interleaved reasoning mechanism that alternates between textual reasoning and spatial grounding, enabling physically grounded prediction. This design yields superior performance on tasks requiring precise localization, counting, and other embodied perceptual reasoning capabilities. 
RynnBrain-Nav demonstrates that adopting RynnBrain as a backbone substantially elevates performance ceilings across various embodied tasks.
RynnBrain‑Plan validates the effectiveness of the fine‑grained manipulation‑planning paradigm that alternates between textual reasoning and localization. 
Finally, RynnBrain-VLA shows that embodiment-agnostic foundational pretraining under the RynnBrain paradigm benefits downstream VLA models that directly predict low-level actions.



A fundamental bottleneck for embodied foundation models is the scarcity of high-quality training data. 
We observe that more realistic and diverse data can substantially enrich and deepen RynnBrain’s capabilities in real-world scenarios. 
To this end, we design dedicated data pipelines tailored to key competencies, including OCR, spatio-temporal localization, action planning, and physically grounded reasoning. 
Importantly, our data construction framework strategically leverages the priors of pretrained foundation models, introducing human supervision only at critical decision points. This human–model collaborative data flywheel improves annotation efficiency and data quality under constrained labeling budgets, enabling the training corpus to scale over 20 million samples.




We extensively evaluate the proposed RynnBrain models in multiple dimensions. Also, recognizing that existing open-source benchmarks inadequately assess fine-grained understanding and spatio-temporal localization, we introduce \textbf{RynnBrain-Bench}, a curated benchmark with carefully filtered and manually verified annotations to ensure robustness and reliability. 
Across 28 benchmarks, RynnBrain demonstrates strong egocentric cognition, including spatial and temporal understanding, OCR, and robot question answering, as well as diverse localization capabilities spanning objects, areas, affordances, and trajectories. Meanwhile, it retains competitive general-purpose visual understanding and instruction-following capabilities.


We further evaluate four post-trained variants across distinct embodied domains: spatio-temporal reasoning (RynnBrain-CoP), vision-and-language navigation (RynnBrain-Nav), manipulation planning (RynnBrain-Plan), and vision-language-action modeling (RynnBrain-VLA). The interleaved grounding–reasoning paradigm of RynnBrain-CoP improves performance on complex spatio-temporal tasks (e.g., trajectory prediction) by approximately 7\%. On the R2R~\cite{anderson2018vision} and RxR~\cite{ku2020room} benchmarks, RynnBrain-Nav achieves state-of-the-art results and consistently surpasses Qwen3-VL-based counterparts across model scales.
For manipulation planning, RynnBrain-Plan adopts two online evaluation protocols, VLMs-UMI and VLMs-VLA. VLMs-UMI directly measures the accuracy and efficiency of high-level planning, while The VLMs-VLA framework evaluates how RynnBrain’s physics-aware, spatially explicit plans enhance downstream VLA execution, thereby strengthening the robustness of the hierarchical embodied architecture.
In high-complexity grasping scenarios, RynnBrain-VLA consistently outperforms models fine-tuned from $\pi_{0.5}$~\cite{pi_0.5}, 
indicating that strong scene understanding and embodied grounding form a critical foundation for generalizable VLA systems.

All code, model checkpoints, and benchmarks are publicly released to facilitate reproducibility and further research.
We envision RynnBrain as a foundational step toward physically grounded general intelligence, where unified spatio-temporal reasoning and physics-aware planning enable embodied agents to operate robustly across diverse real-world settings.

\section{Overview}
\subsection{Model Architecture}
\begin{figure*}
    \centering
    \includegraphics[width=\linewidth]{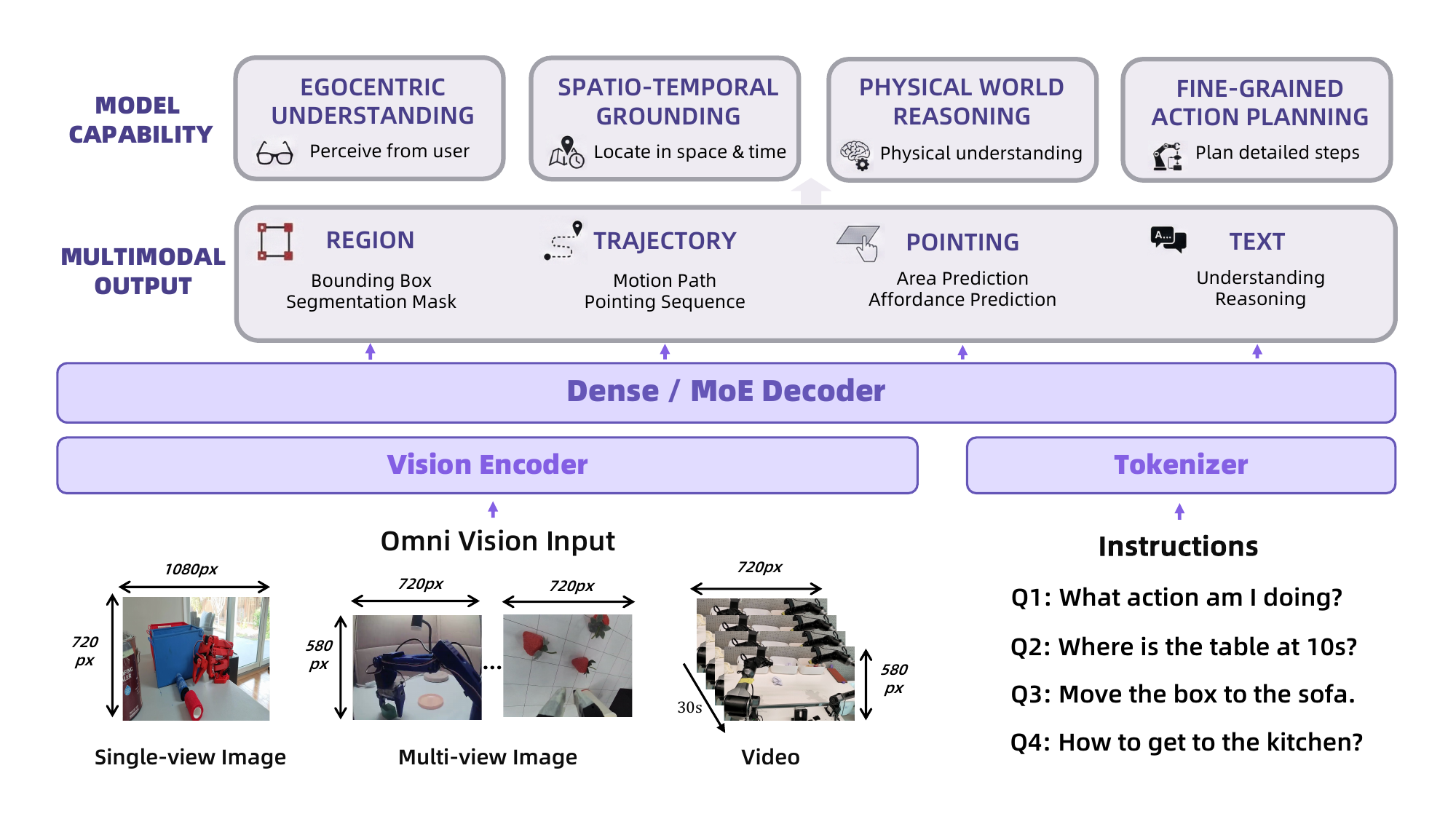}
    \caption{
    Overview of the RynnBrain architecture. RynnBrain processes omni vision inputs, including single view images, multi view images, and videos, together with language instructions. A shared dense or mixture of experts decoder generates aligned multimodal outputs, including text, regions, trajectories, and pointing signals. This unified output space supports egocentric understanding, spatiotemporal grounding, physically grounded reasoning, and fine grained action planning in real world environments.
    }
    \label{fig:model framework}
\end{figure*}

\todo[inline, color=myblue]{Assigned to: Jiayan Guo}
An overview of the RynnBrain architecture is shown in \Cref{fig:model framework}. 
RynnBrain adopts a decoder-only vision–language architecture following the design principles of Qwen3-VL~\cite{qwen3vl}.
It comprises a vision encoder, a vision-language projector, and a large language model~(LLM) backbone initialized from Qwen3-VL variants~(Qwen3-VL-2B/8B/30B-A3B-Instruct). 
In addition, we also employ the techniques of DeepStack~\cite{meng2024deepstack} and Interleaved MRoPE~\cite{huang2025revisiting} to better integrate multimodal information.

\subsection{Infrastructure}
\todo[inline, color=mygray!20]{Assigned to: Kehan Li}

As a general-purpose embodied foundation model, the training data of RynnBrain consists of multiple modalities—including video, image, and text—across a wide range of tasks. These tasks range from short-response tasks, such as localization and spatial perception, to long-form tasks involving detailed multimodal captioning and complex reasoning.
This inherent task diversity results in sequence length distributions characterized by high variance and a pronounced long-tail profile.
Since computational complexity scales with sequence length, a naive distribution of samples across a data parallel (DP) training environment induces a severe straggler effect, where workers assigned heavy workloads become throughput bottlenecks.

To mitigate this, we implement an online load-balancing pipeline. Specifically, we first estimate the sequence lengths of all samples according to pre-computed image sizes and the numbers of text tokens.
During the batch sampling phase of the training process, rather than assigning an equal number of samples to each DP worker, we aggregate all samples across the DP group and redistribute them based on the objective of minimizing the maximum cumulative sequence length within each DP worker.
To solve this redistribution efficiently, we adopt a greedy approximation algorithm that prioritizes longer sequences: we initialize buffers equal to the DP world size, sort sequences in descending order of length, and iteratively assign each to the buffer with the smallest current total length. 
This process is executed during data prefetching; under the Single Program, Multiple Data (SPMD) framework, stable sorting ensures that global data distribution remains consistent across all workers.
This fast and dynamic allocation prevents training stalls while maintaining flexibility, eliminating the need for costly data pre-processing when hyperparameters or datasets change.

To maintain convergence stability after sample redistribution, the global number of tokens is required under the traditional per-token loss formulation:
\begin{equation}
\mathcal{L} = \frac{1}{\sum_{i=1}^n \sum_{j=1}^{b_i} s_{ij}} \sum_{i=1}^n \sum_{j=1}^{b_i} \sum_{k=1}^{s_{ij}} l_{ijk},
\end{equation}
where $ n $ is the DP world size, $ b_i $ is the local batch size on $ i $-th worker, $ s_{ij} $ is the sequence length of the $ j $-th sequence, and $ l_{ijk} $ is the per-token loss.
However, calculating the global token count (the denominator) requires an additional all-gather operation across the DP group, which introduces synchronization overhead and reduces training efficiency.
To circumvent this, we adopt a per-sample loss reduction strategy:
\begin{equation}
\mathcal{L} = \frac{1}{b} \sum_{i=1}^n \sum_{j=1}^{b_i} \frac{1}{s_{ij}} \sum_{k=1}^{s_{ij}} l_{ijk},
\end{equation}
where $ b $ is the global batch size.
Since $ b $ is a constant known to each worker, this strategy eliminates the need for extra communication and improves efficiency.
The holistic approach doubles training efficiency while preserving model stability and convergence properties.


To accommodate the models within the memory constraints of a single GPU, we employ the ZeRO-1 optimizer~\cite{deepspeed} and per-block gradient checkpointing for training RynnBrain-2B and RynnBrain-8B.
Considering the large memory consumption of the logits, we selectively filter out tokens that do not require loss calculation—such as multimodal tokens—during the forward pass of the output head.
For the larger RynnBrain-30B-A3B model, we employ the ZeRO-2 optimizer~\cite{deepspeed} and expert parallel (EP) with a world size of 2 to partition and fit the model in a single GPU.
To optimize computational throughput, we implement the grouped linear operation for MoE layers with packed inputs and weights based on the kernel templates from NVIDIA CUTLASS~\footnote{\url{https://github.com/NVIDIA/cutlass}}.
Cross-GPU token dispatching for EP is facilitated via DeepEP~\cite{deepseekv3}.
For broad accessibility and extensibility, our training and inference frameworks are basedon the HuggingFace Transformers~\cite{transformers} library and have been released as open-source.

\section{Physics-Aware Spatio-temporal Pretraining}
\noindent
Enabling generalizable robots to interact naturally with real-world environments requires two fundamental capabilities:
(1) Spatio-temporal Memory: Through historical visual memory, the robot must establish multi-dimensional representations encompassing space, location, events, trajectories, and e.g., thereby enabling adaptation to complex and dynamic environments.
(2) Physical World Grounding: All robotic cognitive processes must be fundamentally rooted in the objective reality of the physical world.
This chapter primarily introduces the pretraining methodology of RynnBrain, which is explicitly guided by these two insights.

\subsection{Training Recipe}
\todo[inline, color=myblue]{Assigned to: Jiayan Guo}

To equip RynnBrain with spatio-temporal memory and physical grounding, we adopt a unified pretraining framework that maps multimodal inputs into a shared semantic representation space. The training recipe is structured around two core components: a unified input–output representation and a physics-aware optimization strategy.

\paragraph{Unified Spatio-temporal Representation.}
To support spatio-temporal memory, we treat images and videos as a unified visual modality. Formally, a visual input $\mathbf{V}$ is represented as a sequence of frames $\{I_t\}_{t=1}^T$, where $T=1$ for static images and $T>1$ for videos. For videos, frames are uniformly sampled to preserve temporal continuity. Each frame is encoded into visual tokens and augmented with temporal positional embeddings to encode frame order. This representation enables RynnBrain to capture temporal dependencies, motion patterns, and trajectory dynamics across extended visual sequences.

\paragraph{Physically Grounded Output Space.}
To ensure physical world grounding, we explicitly structure the output space to bridge high-level reasoning and low-level execution. Unlike conventional vision–language models that treat spatial quantities as free-form text, we introduce discrete coordinate tokens to represent physical locations. All spatial entities, including bounding boxes $\mathcal{B}$, points $\mathcal{P}$, and trajectory waypoints $\mathcal{T}$, are normalized to the range $[0,1000]$ and encoded as integer tokens. This discretization converts continuous spatial prediction into a classification problem, allowing the model to generate precise and physically meaningful spatial outputs using the same autoregressive mechanism as language generation.

\paragraph{Optimization.}
RynnBrain is trained end to end using a standard next-token prediction objective. The training loss is defined as:
\begin{equation}
    \mathcal{L} = - \sum_{i=1}^{L} \log P\left( y_i \mid y_{<i}, \mathbf{V}, \mathbf{\Theta} \right),
\end{equation}
where $\mathbf{V}$ denotes the visual input, $\mathbf{y}$ is the mixed sequence of textual and coordinate tokens, and $\mathbf{\Theta}$ represents the model parameters. Optimization hyperparameters are adjusted across model scales based on pilot experiments conducted on a representative subset of the pretraining data. Detailed training configurations are reported in \Cref{tab:hyperparams}.

\begin{table}[th]
    \centering
    \caption{Hyperparameters of the pretraining stage for RynnBrain model series.}
    \label{tab:hyperparams}
    \resizebox{\textwidth}{!}{
    \begin{tabular}{l|ccc}
        \hline
        \textbf{Parameter} & \textbf{RynnBrain-2B} & \textbf{RynnBrain-8B} & \textbf{RynnBrain-30B-A3B} \\
        \hline
        
        Base Model & Qwen3-VL-2B-Instruct & Qwen3-VL-8B-Instruct & Qwen3-VL-30B-A3B-Instruct \\
        Optimizer & AdamW & AdamW & AdamW \\
        Learning Rate & $5e^{-6}$ & $2e^{-6}$ & $2e^{-6}$ \\
        Learning Rate Vision & $1e^{-6}$ & $2e^{-6}$ & $2e^{-6}$ \\
        Global Batch Size & 512 & 1024 & 1024 \\
        Warmup Ratio & 0.03 & 0.03 & 0.03 \\
        \hline
    \end{tabular}
    }
\end{table}


\subsection{Pretraining Data}
\label{sec:pretraindata}
\Cref{tab:data-mixture} summarizes the data sources and corresponding data volumes used for pretraining RynnBrain. Below, we describe each dataset grouped by category.
\begin{table}[t]
    \centering
    \caption{Pretraining data mixture statistics for RynnBrain}
    \label{tab:data-mixture}
    
    \begin{tabularx}{\textwidth}{>{\raggedright\arraybackslash}p{2.2cm}|>{\raggedright\arraybackslash}p{3.7cm}|>{\raggedright\arraybackslash}X|p{1cm}}
        \hline
        \textbf{Category} & \textbf{Sub-Task} & \textbf{Data Sources} & \textbf{Samples (M)} \\
        \hline
        
         \multirow{4.5}{*}{\textbf{General MLLM}} & \multirow{4.5}{*}{General} & LLaVA-OV-SI~\cite{llava-OV}, 
        LLaVA-Video~\cite{llava-video}, 
        ShareGPT-4o-video~\cite{sharegpt4video}, VideoGPT-plus~\cite{videogpt+}, FineVideo~\cite{FineVideo}, CinePile~\cite{cinepile}, ActivityNet~\cite{activitynet}, YouCook2~\cite{youcook2}, LLaVA-SFT~\cite{llava} & \multirow{5}{*}{4.80} \\
        \hline
        
        \multirow{11}{*}{\textbf{Cognition}} 
         & Object Understanding & RynnBrain-Object, RefCOCO~\cite{yu2016modeling}, Google Refexp~\cite{mao2016generation}, Osprey-724K~\cite{yuan2024osprey}, DAM~\cite{lian2025describe}, VideoRefer-700k~\cite{yuan2025videorefer}  & 1.10 \\ 
         & Spatial Understanding & Sensenova-SI-800K~\cite{cai2025scaling}, VSI-590k~\cite{yang2025cambrian}, VLM-3R~\cite{fan2025vlm}, RynnBrain-Spatial & 2.50 \\ 
         & Counting & RynnBrain-Counting, Molmo2~\cite{clark2026molmo2} & 0.30 \\ 
         & OCR & RynnBrain-OCR & 1.00 \\
        
        & Egocentric Task Understanding & EgoRe-5M~\cite{pei2025egothinker}, Egotaskqa~\cite{jia2022egotaskqa}, Env-QA~\cite{Gao_2021_ICCV}, QAEgo4d~\cite{grauman2022ego4d},
        RoboVQA~\cite{sermanet2024robovqa}, Robo2vlm~\cite{chen2025robo2vlm}, 
        ShareRobot~\cite{ji2025robobrain} & 2.77 \\
        
        \hline
        
        \multirow{7}{*}{\textbf{Localization}} 
         & Object Localization & ADE20K~\cite{ade20k}, COCOStuff~\cite{cocostuff}, Mapillary~\cite{mapillary}, PACO-LVIS~\cite{paco}, PASCAL-Part~\cite{pascal-part}, VG~\cite{krishna2017visual} RoboAfford-Object~\cite{hao2025roboafford++}, RynnBrain-Grounding & 1.20 \\ 
         & Area Localization & RefSpatial~\cite{zhou2025roborefer}, RoboAfford-Area~\cite{hao2025roboafford++}, Molmo2~\cite{clark2026molmo2}, RynnBrain-Area & 3.37 \\ 
         & Affordance Localization & RynnBrain-Affordance, RoboAfford-Affordance~\cite{hao2025roboafford++} & 1.13 \\ 
         & Trajectory Prediction & RynnBrain-Trajectory, FSD~\cite{yuan2025seeingdoingbridgingreasoning} & 0.56 \\ 
         & Grasp Pose Prediction & Grasp-Anything~\cite{vuong2024grasp} & 1.00 \\
        \hline
        
        \multirow{2}{*}{\textbf{Planning}} 
         & Manipulation & AgibotWorld~\cite{contributors2024agibotworldrepo}, 
         Open X-Embodiment~\cite{open_x_embodiment_rt_x_2023}, RynnBrain-Planning & 0.16 \\
        \hline
        \hline
\multicolumn{1}{l}{\textbf{Total}} & \multicolumn{1}{c}{} &  & \multicolumn{1}{c}{\textbf{19.89}} \\        \hline
\end{tabularx}
\end{table}
\subsubsection{General MLLM Data}
\todo[inline, color=myred!20]{Assigned to: Yuqian Yuan}
To retain broad multimodal understanding, we construct a general-purpose MLLM pretraining corpus spanning both images and videos across diverse domains. The corpus aggregates publicly available datasets, including LLaVA-OV-SI~\cite{llava-OV}, LLaVA-Video~\cite{llava-video}, ShareGPT-4o-video~\cite{sharegpt4video}, VideoGPT-plus~\cite{videogpt+}, FineVideo~\cite{FineVideo}, CinePile~\cite{cinepile}, ActivityNet~\cite{activitynet}, YouCook2~\cite{youcook2}, LLaVA-SFT~\cite{llava}, and VideoLLaMA 3~\cite{zhang2025videollama}. 
Together, these datasets support open-vocabulary object recognition, conversational video understanding, long-horizon temporal reasoning, and image–text supervision. In total, the corpus comprises 4.8M samples.

\subsubsection{Multi-Dimensional Cognition Data}

\paragraph{Object Understanding.}
The object understanding dataset is designed to enhance fine-grained object recognition and object-centric reasoning. Each sample focuses on a specific object annotated with a bounding box in a single frame, formatted as \texttt{<object> <frame n>: (coordinates) </object>}, with questions conditioned on the indicated object. The dataset covers object attributes such as category, color, shape, function, spatial position, and related properties.

We combine publicly available datasets~\cite{yu2016modeling, mao2016generation, yuan2024osprey, lian2025describe, yuan2025videorefer} with self-collected egocentric data, yielding over 1.1M samples. For the egocentric subset, we construct an object-centric QA generation pipeline on indoor videos. Objects are first identified using Qwen2.5-VL~\cite{qwen2.5vl}, detected in key frames with Grounding DINO 1.5~\cite{ren2024grounding}, and segmented and tracked using SAM2~\cite{ravi2024sam}. To reduce redundancy, we limit each video to at most two instances per object category. Object-centric QA pairs are then generated using Qwen2.5-VL and manually filtered for quality, resulting in 712K high-quality QA samples.

\paragraph{Spatial Understanding.}
Spatial reasoning is critical for embodied tasks such as navigation and manipulation, yet remains a weakness of many existing VLMs. To address this limitation, we curate over 2.5M spatial instruction samples spanning two categories: general spatial understanding and fine-grained object-centric spatial reasoning.

General spatial understanding data are sourced from publicly available datasets, including Sensenova-SI-800K~\cite{cai2025scaling}, VLM-3R~\cite{fan2025vlm}, and VSI-590K~\cite{yang2025cambrian}. For fine-grained spatial annotations, we process self-collected indoor images and videos using MASt3R-SLAM~\cite{murai2025mast3r}, which reconstructs 3D point clouds and estimates camera extrinsics from RGB video. Instance-level segmentations are projected into the reconstructed 3D space, and the point cloud is realigned using RANSAC~\cite{fischler1981random} to detect the ground plane and enforce a gravity-aligned world coordinate system.

Based on these calibrated 3D scenes, we generate spatial QA pairs requiring reasoning about metric distances, relative positions, heights, and other 3D relationships. QA generation follows a template-based scheme, where missing attributes are computed directly from the underlying geometry. This process yields 855K video-based and 272K image-based spatial QA samples.

\paragraph{Counting.}
\todo[inline, color=myred!20]{Assigned to: Yuqian Yuan}
The counting dataset is designed to improve robust estimation of object quantities in complex visual scenes. We combine publicly available data with egocentric indoor videos. The public component consists of the Molmo2 counting subset~\cite{clark2026molmo2}, comprising 222K samples with diverse scenes and reliable annotations. To incorporate embodied perspectives, we further curate 42K counting QA pairs from self-collected egocentric videos. All annotations are manually verified to ensure accuracy and consistency.

\paragraph{OCR.}
\todo[inline, color=myyellow!20]{Assigned to: Sicong Leng}
The OCR dataset equips the model with scene text recognition and grounding capabilities essential for text-rich embodied environments. We construct approximately 1M OCR QA samples from egocentric videos sourced from Ego4D~\cite{grauman2022ego4d}, Charades-Ego~\cite{sigurdsson2018charades}, and EPIC-KITCHENS~\cite{damen2018scaling}. Scene text is detected using GoMatching~\cite{he2024gomatching, he2025gomatching++}, and videos are segmented based on text appearance patterns into clips of 3 to 15 seconds, yielding 85,324 text-containing segments.

For each segment, human annotators label the first appearance frame, the clearest frame, text transcription, and bounding polygons. QA pairs are generated using two complementary strategies: (i) GPT-5.2~\cite{openai2025gpt52} produces goal-oriented, first-person questions grounded in practical text understanding, yielding 256K contextual QA samples; (ii) template-based generation produces structured questions covering text reading, temporal localization, verification, and multiple-choice recognition, yielding 722K samples. GPT-generated questions are filtered to ensure visual perception is required.

The OCR dataset is provided in two formats: normal video QA (893K samples), where the model predicts textual answers from video input, and area prediction QA (85K samples), where the model outputs frame indices and normalized bounding coordinates.

\paragraph{Egocentric Task Understanding}
\todo[inline, color=myred!20]{Assigned to: Yuqian Yuan}
To support broad egocentric task comprehension, we construct an egocentric task understanding dataset comprising 2.77M video–text pairs. The dataset aggregates publicly available resources, including Env-QA~\cite{Gao_2021_ICCV}, EgoTaskQA~\cite{jia2022egotaskqa}, RoboVQA~\cite{sermanet2024robovqa}, EgoRe-5M~\cite{pei2025egothinker}, QAEgo4D~\cite{patel2025advancing}, Robo2VLM~\cite{chen2025robo2vlm}, and ShareRobot~\cite{ji2025robobrain}. Videos shorter than 3 seconds are excluded to ensure sufficient temporal context for task-level reasoning.



\subsubsection{Spatio-Temporal Location Data}
\paragraph{Object Location.}
\todo[inline, color=myred!20]{Assigned to: Yuqian Yuan}
Object localization enables the model to interpret language instructions and identify target objects in images and videos. Each sample is represented as $(\mathbf{V}, Q, \mathcal{B}, t)$, where $\mathbf{V}=\{I_t\}_{t=1}^T$ denotes a sequence of $T$ frames ($T=1$ for static images), $Q$ is a textual query describing the target object, $\mathcal{B}=\{(x_0,y_0,x_1,y_1)\}$ is the bounding box of the target with normalized coordinates in $[0,1000]$, and $t$ denotes the key frame where the object is most clearly observed.

We aggregate 900K samples from publicly available grounding datasets, including ADE20K~\cite{ade20k}, COCO~\cite{lin2014microsoft}, Mapillary~\cite{mapillary}, PACO-LVIS~\cite{paco}, PASCAL-Part~\cite{pascal-part}, VG~\cite{krishna2017visual}, and RoboAfford++~\cite{hao2025roboafford++}. To strengthen egocentric localization, we further construct 300K egocentric samples using the same segmentation pipeline as object understanding. Referring expressions are generated using Qwen3~\cite{qwen3}, including simple expressions based on category or position and situational expressions that require task-level inference. All samples are manually filtered for quality.

\paragraph{Area Location.}
\todo[inline, color=mygreen!20]{Assigned to: Zhikai Wang}
Area localization equips the model to identify non-object regions, such as surfaces, empty spaces, or functional areas, in images and videos. Each sample is represented as $(\mathbf{V}, Q, \mathcal{P}, t)$, where $\mathcal{P}=\{(x_i,y_i)\}_{i=1}^n$ denotes a set of normalized points indicating the target area, and $t$ is the keyframe index.

The dataset is constructed from multiple sources. We annotate 6K egocentric house-touring video segments using LLM–generated instructions with human-selected point annotations. To enhance temporal coverage, we incorporate 222K video samples from Molmo2-VideoPoint~\cite{clark2026molmo2}. For static scenes, we curate 448K image–area samples from indoor images using a similar pipeline. Additionally, we include 2.2M image-based samples from RoboAfford++~\cite{hao2025roboafford++} and RefSpatial~\cite{zhou2025roborefer} to increase domain diversity.

\paragraph{Affordance Location.}
\todo[inline, color=mygreen!20]{Assigned to: Zhikai Wang}
Affordance localization focuses on identifying actionable points, such as handles, buttons, or interaction hotspots, on objects or surfaces. Each sample is represented as $(\mathbf{V}, Q, p, t)$, where $p=(x,y)$ is a normalized affordance point and $t$ denotes the key frame index where the affordance is most relevant..

We follow a construction pipeline similar to area localization. For spatiotemporal data, we annotate 6K video segments with LLM–generated instructions and human-labeled affordance points. For static images, we derive 476K affordance samples from 500K indoor images. To improve generalization, we further include 260K affordance samples from RoboAfford++~\cite{hao2025roboafford++}, focusing on actionable interactions.



\paragraph{Trajectory Location}
\todo[inline, color=mygreen!20]{Assigned to: Zhikai Wang}
Trajectory localization trains the model to predict plausible two-dimensional manipulation trajectories for object interaction tasks. Each sample is represented as $(\mathbf{V}, Q, \mathcal{T}, t_s)$, where $\mathcal{T}=\{(x_i,y_i)\}_{i=1}^m$ is an ordered set of up to 10 normalized trajectory points, and $t_s$ denotes the starting frame.

We construct 6K spatiotemporal samples with LLM–generated instructions and human-annotated trajectories, emphasizing cross-frame reasoning. For static images, we generate 507K image–trajectory samples from indoor scenes. To further diversify manipulation scenarios, we include 13K trajectory samples from FSD~\cite{yuan2025seeingdoingbridgingreasoning}.



\paragraph{Grasp Pose Location}
\todo[inline, color=myyellow!20]{Assigned to: Sicong Leng}
The grasp pose location dataset equips the model with the ability to predict precise robotic grasp poses for target objects. Each sample is represented as $(I, Q, \mathcal{G})$, where $I$ denotes a single RGB image, $Q$ is a textual query specifying the target object and grasping task, and $\mathcal{G}=\{(x_i,y_i)\}_{i=1}^4$ denotes four ordered corner points defining an oriented grasp rectangle.

We construct this dataset from Grasp-Anything~\cite{vuong2024grasp}, which provides grasp annotations for everyday objects in tabletop scenes using oriented rectangles parameterized by center $(c_x,c_y)$, dimensions $(w,h)$, and rotation angle $\theta$. We process approximately 995K images at $416\times416$ resolution, each containing one or more annotated grasp candidates. For each object, we select the highest-scoring grasp and convert the parameterized representation into four corner points via rotation. This representation explicitly captures grasp orientation and gripper alignment, supporting spatially precise manipulation planning.

To promote linguistic diversity, we generate instruction prompts using a weighted template strategy: 40\% object-centric prompts, 30\% scene-aware prompts incorporating scene descriptions, and 30\% task-oriented prompts emphasizing manipulation intent. Grasp pose outputs are expressed using multiple concise response templates to improve robustness to varied linguistic formulations.

Following this pipeline, we construct a static-image grasp pose dataset comprising 1.3M training samples derived from 945K images, with an average of 1.44 samples per image. This dataset enables RynnBrain to learn orientation-aware and spatially grounded grasp pose prediction, a key capability for robotic manipulation.

\subsubsection{Physics-Aware Planning Data}



\todo[inline, color=mybrown!20]{Assigned to: Yunxuan Mao}


To support precise manipulation planning, we design a structured planning data schema. Following Hi Robot~\cite{shi2025hirobotopenendedinstruction}, we adopt atomic actions as the minimal units of planning. Long-horizon tasks are decomposed into temporally ordered sub-tasks using an in-house model and subsequently verified by human annotators. 

To enable fine-grained spatial grounding, each sub-task is annotated with a unified grounding schema that includes target object bounding boxes, placement area points, and affordance points. Formally, each training sample is represented as $(\mathbf{V}, Q, \mathcal{M})$, where $\mathbf{V}=\{I_t\}_{t=1}^T$ denotes the visual context preceding the current step, $Q$ is a high-level task instruction (e.g., ``Please help me tidy up the sink.''), and $\mathcal{M}$ denotes the current sub-task plan, expressed as a mixed sequence of textual tokens and grounding annotations (bounding boxes $\mathcal{B}$, area points $\mathcal{P}$, and affordance point $p$).

We incorporate publicly available datasets, including AgibotWorld Alpha~\cite{contributors2024agibotworldrepo} and Open X-Embodiment~\cite{open_x_embodiment_rt_x_2023}, formatted as single-turn planning dialogues. To strengthen physical grounding, we further augment these data with spatial annotations on randomly sampled frames. This design enables RynnBrain to integrate object, region, and affordance information directly into planning outputs, providing downstream manipulation policies with precise and physically grounded guidance.

\section{Physically Grounded Chain-of-Point Reasoning}
\noindent

%
Most existing multimodal reasoning models~\cite{Video-r1, deepseek-vl,leng2025mmr1} rely on purely textual reasoning paradigms. Although several approaches~\cite{taco, deepeyes, refocus, yang2025longvt} incorporate auxiliary tools such as region zooming to alleviate visual recognition challenges, their reasoning processes remain largely detached from physical spatial structure, limiting generalization beyond narrowly defined tasks. Alternative methods that explore visual imagination during reasoning~\cite{PARM, VPRL, Got-r1} further suffer from hallucinated visual content, undermining physical consistency.

For embodied agents operating in real-world environments, reasoning must be grounded in observable physical evidence. To this end, we introduce \emph{Chain-of-Point} (\textbf{CoP}) reasoning in RynnBrain, an interleaved reasoning paradigm that integrates explicit spatial grounding with textual inference over egocentric video streams. By anchoring intermediate reasoning steps to concrete spatial references, CoP bridges language-based cognition and physical perception, enabling reasoning that remains consistent with the underlying environment. This section presents the design and explorations of CoP reasoning for physically grounded embodied intelligence.


\subsection{Cold-Start Supervised Fine-Tuning}
\subsubsection{Training Recipe}
\todo[inline, color=myblue]{Assigned to: Jiayan Guo}

The training of the CoP reasoning model, i.e., RynnBrain-CoP, begins with the pretrained RynnBrain model, which establishes a strong foundation in general embodied understanding. We perform full-parameter supervised fine-tuning using the AdamW optimizer with a cosine learning rate schedule. We set the peak learning rate to $1 \times 10^{-5}$ for the language model and projector, and $2 \times 10^{-6}$ for the vision encoder, with a 3\% warmup period. The model is trained for 1 epoch with a global batch size of 128. To effectively process long-horizon egocentric videos, we sample frames at 2 FPS (up to 2048 frames) and set the maximum context length to 16,384 tokens. We utilize DeepSpeed ZeRO-1 to optimize memory efficiency during training.

\subsubsection{Data}
\todo[inline, color=mygreen!20]{Assigned to: Zhikai Wang}
To develop the model's CoP reasoning capability, we construct a specialized dataset that explicitly interleaves textual reasoning with visual grounding. This process is based on the core spatio-temporal location datasets (Area, Affordance, and Trajectory Location). Each sample is enriched with a ``Thinking'' field that bridges high-level task understanding with low-level spatial localization.

The data generation follows a structured pipeline: First, given the original task instruction and video frames, we use Qwen3-VL-235B to pre-generate a step-by-step textual reasoning chain. This chain includes key reasoning steps and explicitly marks potential entities (e.g., objects or areas) using square brackets (e.g., [white flower-patterned wallpaper]). These entities are candidates for visual grounding. Next, an in-house model is employed to classify each marked entity as either ``area'' or ``object'' based on the textual context. Finally, human annotators review the reasoning chain and entity classifications. For each identified entity, they select the most relevant and clear frame from the video sequence and perform precise annotation: for entities classified as ``area'', they annotate a set of representative points; for those classified as ``object'', they annotate a 2D bounding box. The grounding results are then inserted back into the reasoning text in the structured format \texttt{<object/area> <frame n>: ...; (coordinates) </object/area>}, creating a seamless interleaving of textual reasoning and spatial grounding.

This process results in a CoT-style dataset where the model's internal thinking process is not merely abstract, but is continually anchored to specific visual evidence in the physical space. Formally, a sample extends the base tuple to $(V, Q, \mathcal{P}_{final}, t_s, R)$, where $R$ represents this interleaved reasoning chain. This dataset is fundamental for training RynnBrain-CoP, enabling RynnBrain to perform transparent, grounded, and hallucination-resistant reasoning essential for reliable operation in embodied scenarios.



\subsection{Reinforcement Learning}
\subsubsection{Training Recipe}
\todo[inline, color=myblue]{Assigned to: Jiayan Guo}


We employ Group Relative Policy Optimization (GRPO)~\cite{shao2024deepseekmath} to align the model with physically grounded reasoning tasks. Unlike standard PPO~\cite{schulman2017proximal} which requires a value function (critic) to estimate the advantages, GRPO estimates the baseline from the group scores of multiple sampled outputs generated from the same prompt. This significantly reduces memory usage and training complexity.

Formally, for each query $q$, we sample a group of $G$ outputs $\{o_1, o_2, ..., o_G\}$ from the old policy $\pi_{\theta_{old}}$. The optimization objective is defined as follows:

\begin{equation}
    \mathcal{J}_{\text{GRPO}}(\theta) = \mathbb{E}\left[ \frac{1}{G} \sum_{i=1}^G \left( \min \left( \rho_i A_i, \text{clip}(\rho_i, 1-\epsilon, 1+\epsilon) A_i \right) - \beta \mathbb{D}_{KL}(\pi_\theta(o_i|q) || \pi_{\text{ref}}(o_i|q)) \right) \right]
\end{equation}

where $\rho_i = \frac{\pi_\theta(o_i|q)}{\pi_{\theta_{old}}(o_i|q)}$ is the importance sampling ratio, and $\beta$ is the coefficient for the KL divergence penalty with respect to the reference model $\pi_{\text{ref}}$. The advantage $A_i$ for each output is computed by normalizing the rewards within the group:
\begin{equation}
    A_i = \frac{r_i - \text{mean}(\{r_1, ..., r_G\})}{\text{std}(\{r_1, ..., r_G\}) + \epsilon}
\end{equation}

The training is initialized from our cold-start SFT model. We utilize the SGLang~\cite{zheng2024sglang} inference engine for efficient rollout generation with a group size of $G=5$. The training runs for 10 epochs with a batch size of 128. We optimize the policy using a cosine learning rate schedule starting at $2\times 10^{-6}$ with a 3\% warmup. To ensure stability, we set the clipping range $\epsilon$ to $[0.2, 0.28]$ and the KL coefficient $\beta=0.02$. The maximum sequence length is set to 16,384 tokens to accommodate long-context egocentric video reasoning.

\subsubsection{Reward Design}
We design task-specific rule-based reward functions to strictly anchor the model's reasoning in the physical world. All spatial coordinates are normalized to the unit interval $[0, 1]$ prior to reward computation.

\paragraph{Trajectory.}
The trajectory reward evaluates the shape and sequential alignment of the predicted path. First, both the predicted sequence $\mathcal{P} = (p_1, \dots, p_M)$ and the ground truth sequence $\mathcal{G} = (g_1, \dots, g_N)$ are resampled to have the same number of points uniformly spaced by arc length. We then calculate the Discrete Fréchet Distance (DFD), defined recursively. Let $c(i, j)$ be the coupling distance between prefixes $p_{1:i}$ and $g_{1:j}$:
\begin{equation}
    c(i, j) = \max \left( \|p_i - g_j\|_2, \min \big( c(i-1, j), c(i, j-1), c(i-1, j-1) \big) \right)
\end{equation}
with $c(0, 0) = \|p_1 - g_1\|_2$. The final distance is $D_{F} = c(M, N)$. The reward decays exponentially with this distance:
\begin{equation}
\label{eq:dfd}
    r_{\text{traj}} = \exp(-\lambda_{\text{traj}} \cdot D_{F})
\end{equation}

\paragraph{Affordance.}
For affordance, we measure the set similarity between predicted interaction points $\mathcal{P}$ and ground truth points $\mathcal{G}$ using the Bidirectional Mean Euclidean Distance, a variant of the Chamfer distance. This metric jointly captures precision, by penalizing invalid predictions, and recall, by measuring coverage of all annotated affordance regions:
\begin{equation}
    D_{\text{bidir}}(\mathcal{P}, \mathcal{G}) = \frac{1}{2} \left( \frac{1}{|\mathcal{P}|} \sum_{p \in \mathcal{P}} \min_{g \in \mathcal{G}} \|p - g\|_2 + \frac{1}{|\mathcal{G}|} \sum_{g \in \mathcal{G}} \min_{p \in \mathcal{P}} \|p - g\|_2 \right)
\end{equation}
The reward is defined as $r_{\text{aff}} = \exp(-\lambda_{\text{aff}} \cdot D_{\text{bidir}})$.

\paragraph{Area.}
For area identification, we treat the task as a point-retrieval problem within a valid polygon. Let $S_{\mathcal{G}}$ denote the geometric region defined by the ground truth polygon. The reward is the strict accuracy of the generated points $\mathcal{P}$:
\begin{equation}
    \label{eq:area}
    r_{\text{area}} = \frac{1}{|\mathcal{P}|} \sum_{p \in \mathcal{P}} \mathbb{I}(p \in S_{\mathcal{G}})
\end{equation}
where $\mathbb{I}(\cdot)$ is the indicator function.



\subsubsection{Reinforcement Learning Data}
\todo[inline, color=myorange!20]{Assigned to: Minghao Zhu}
To support efficient and high-quality policy exploration, we construct a curated reinforcement learning dataset based on the spatiotemporal localization data used during pretraining, covering area, affordance, and trajectory tasks. These tasks provide essential supervision for visual evidence localization and physics-aware reasoning.

We apply a difficulty-aware filtering strategy to remove trivial samples that do not require grounded reasoning, as well as excessively noisy or ambiguous cases. Each candidate sample is scored by a pretrained SFT model using the evaluation metrics described in Section~\ref{sec:evaluation:rynnbrain_bench}, and only samples of intermediate difficulty (scores between 40 and 80) are retained. To further improve temporal localization, we additionally include a subset of failure cases in which the SFT model incorrectly selects key frames.

This refinement process yields a high-quality dataset of 30K training samples. By constraining exploration within a structured and physically grounded regime, the dataset reduces hallucinations and promotes more reliable reasoning during reinforcement learning.



\section{Post-training for Embodied Tasks}
\subsection{Vision-Language Navigation}
\todo[inline, color=mypurple!20]{Assigned to: Jiangpin Liu}





The RynnBrain foundation model is pretrained to enhance its general understanding capabilities. To validate the benefits of this pretraining for the task of Vision-Language Navigation (VLN), we subsequently fine-tune the model on a VLN dataset. The fine-tuned RynnBrain model, namely \textbf{RynnBrain-Nav} is then deployed as an agent to perform navigation tasks.

\textbf{Problem Formulation.} In the VLN task, an embodied agent is tasked with interpreting a natural language instruction $Q$. At time $t$, based on a sequence of visual observations $O = \{o_0, o_1, \dots, o_t\}$ and language instruction $Q$, the agent must generate a corresponding action  $a_{t}$ to follow the instruction and reach the target destination. Each observation $o_i \in \mathbb{R}^{3 \times H \times W}$ is an RGB image from the agent's current perspective. The discrete action space is defined as $\mathcal{A} = \{\uparrow, \leftarrow, \rightarrow, \text{STOP}\}$, representing the low-level movements of moving forward by 30~centimeters, turning left or right by 15~degrees, and halting the episode, respectively.

\textbf{Data Formulation.} We adopt a multi-turn conversational format, which is efficient for both training and inference. Following the methodology of StreamVLN~\cite{wei2025streamvln}, the training data is organized as a sequence of observation-action pairs, $d_i = (o_i, a_i)$. The training objective is to predict the next action $a_i$ based on current visual observation $o_i$ and the preceding conversational history. This formulates each VLN trajectory into an interleaved image-text sequence, represented as:
\begin{equation}
\{
o_0, a_0, o_1, a_1, \dots, o_n, a_n
\}
\end{equation}

\textbf{Data Collection.} We curated a large-scale, navigation-specific training dataset using the Habitat simulator~\cite{savva2019habitat} and using the ground truth action to generate the image-text interleaved VLN dataset. The primary component consists of 450K video clips generated from R2R~\cite{anderson2018vision}, R2R-EnvDrop~\cite{envdrop}, and RxR~\cite{ku2020room} trajectories across 60 Matterport3D (MP3D)~\cite{ramakrishnan2021habitat} environments. To enhance scene diversity and improve generalization, this dataset was augmented with an additional 300K samples from a subset of the ScaleVLN~\cite{scalevln} dataset. We further implement multi-turn DAgger~\cite{dagger} to further collect the data to improve the performance of the model.

\textbf{Fine-tuning Settings.} We perform full-parameter supervised fine-tuning using the AdamW optimizer with a cosine learning rate schedule. We set the peak learning rate to $2 \times 10^{-5}$ for the language model and projector, and $2 \times 10^{-6}$ for the vision encoder, with a 3\% warmup period. The model is trained for 1 epoch with a global batch size of 256. To effectively process long-horizon egocentric videos, we sample frames at 2 FPS (up to 2048 frames) and set the maximum context length to 16,384 tokens. We utilize DeepSpeed ZeRO-1 to optimize memory efficiency during training.

\subsection{Manipulation Planning}
\todo[inline, color=mybrown!20]{Assigned to: Yunxuan Mao}
Since the pretraining corpus incorporates planning-centric data, the foundation model already possesses inherent planning capabilities. However, adapting this capability to complex, long-horizon manipulation tasks requires the model to maintain effective memory.
To address this issue, we utilize a tiny in-house dataset formatted as multi-turn dialogues, where the interaction history functions as an explicit memory buffer to preserve historical reasoning results. This structure enables the model to bridge individual planning steps into a coherent long-horizon strategy. Crucially, to align with this sequential inference, grounding annotations were applied exclusively to the final frame of each dialogue turn, ensuring current decisions are conditioned on both the immediate observation and the accumulated memory. Empirically, we find that this approach is highly data-efficient: fine-tuning with only a few hundred samples is sufficient to endow the model with robust long-horizon planning and generalization capabilities. Further details and quantitative evaluations are provided in~\Cref{sec:experiment}.

\subsection{VLA}
\todo[inline, color=mygray!20]{Assigned to: Kehan Li}

\label{sec:vla}

\noindent \textbf{Model Architecture.}
To bridge the gap between planning and physical execution, we propose \textbf{RynnBrain-VLA}, which translates fine-grained plans into executable robot actions.
We build RynnBrain-VLA upon RynnBrain-2B to utilize the large-scale pretraining on fine-grained object references and precise spatial
\begin{wrapfigure}{r}{0.45\textwidth}
    \centering
    \vspace{-1pt}
    \hspace{-5pt}
    \includegraphics[width=0.45\textwidth]{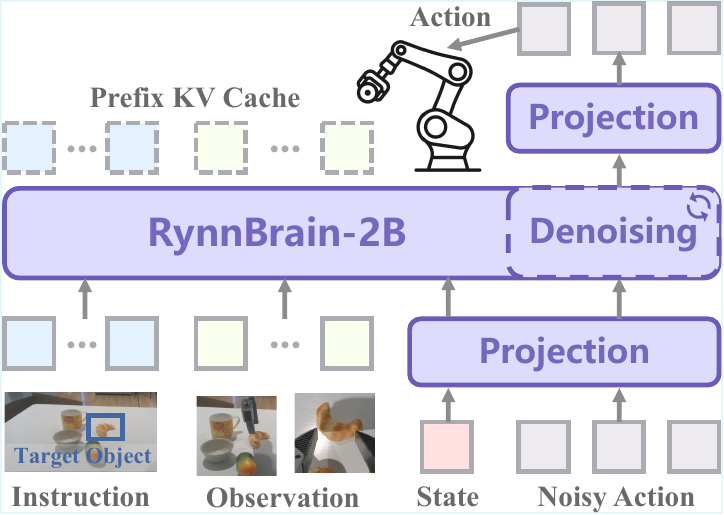}
    \vspace{1pt}
    \caption{RynnBrain-VLA architecture.}
    \vspace{-2pt}
    \label{fig:vla_model}
\end{wrapfigure}
localization while maintaining low inference latency.
The overall model architecture is shown in \Cref{fig:vla_model}.
Generally, we adopt a flow matching framework to predict an action chunk~\cite{pi0} at each step.
The VLM backbone is served as a single-stream Diffusion Transformer (DiT) taking a single packed sequence containing the condition and the noisy actions as input.
To make the VLM compatible with this framework, we add three linear projections to align the dimension of the input noises, input timestamp embeddings, and the output actions to the VLM hidden size.
To preserve the VLM’s inherent instruction-following capabilities, we utilize its native conversation format for organizing the input sequence.
Consistent with the planning model, pointing information is passed in a text-based format, with the task's initial frame prepended to the input sequence.

\fcolorbox{black}{black!5}{
\begin{minipage}{0.975\textwidth}
<|im\_start|>user \\
INSTRUCTION: \\
<start\_frame> \\
Pick the <affordance> ($x$,$y$) </affordance> of the <object> ($x_0$,$y_0$),($x_1$,$y_1$) </object> \\
OBSERVATION: \\
<camera\_1><camera\_2><camera\_3> \\
STATE: \\
<state> \\
What action should the robot take ?<|im\_end|> \\
<|im\_start|>assistant \\
<action>
\end{minipage}
}

Following $ \pi_0 $~\cite{pi0}, actions are positioned at the end of the sequence to enable the KV cache during inference.

\noindent \textbf{Fine-tuning.}
To equip the model with robotic control capabilities, we curated a dataset comprising six pick-and-place tasks involving three distinct objects, collected via manual teleoperation on a Franka Emika arm.
Following data collection, each episode was annotated with its target object or placement location and aligned with the output format of the planning model.
We then fine-tuned the RynnBrain-2B model on this dataset for 60k steps, using a learning rate of 2e-5 and a batch size of 32.
All images were proportionally resized to a short-side dimension of 384 pixels.
Experimental results show that, by leveraging RynnBrain’s embodied understanding and precise localization, this simple adaptation achieves accurate interpretation of point-based instructions and reliable grasping.

\section{Evaluation}
\label{sec:experiment}
\noindent

\subsection{RynnBrain-Bench}
\todo[inline, color=myorange!20]{Assigned to: Minghao Zhu}
\label{sec:evaluation:rynnbrain_bench}
For embodied brains operating in physical reality, the fine-grained spatio-temporal understanding across the entire episodic memory is essential for performing intricate embodied tasks. 
While existing benchmarks primarily focus on either static scene understanding with text-referenced objects~\cite{openeqa, ecbench} or spatial pointing tasks with single-frame input~\cite{ji2025robobrain}, they fall short of adequately evaluating models’ capabilities in this domain.
We introduce \textbf{RynnBrain-Bench}, a high-dimensional evaluation suite designed to holistically benchmark the cognition and localization capabilities of embodied understanding models in complex household environments.
Advancing beyond existing benchmarks, RynnBrain-Bench features a unique emphasis on fine-grained understanding and precise spatio-temporal localization within episodic video sequences.

\subsubsection{Overview}
\begin{figure*}[t]
    \centering
    \includegraphics[width=0.95\linewidth]{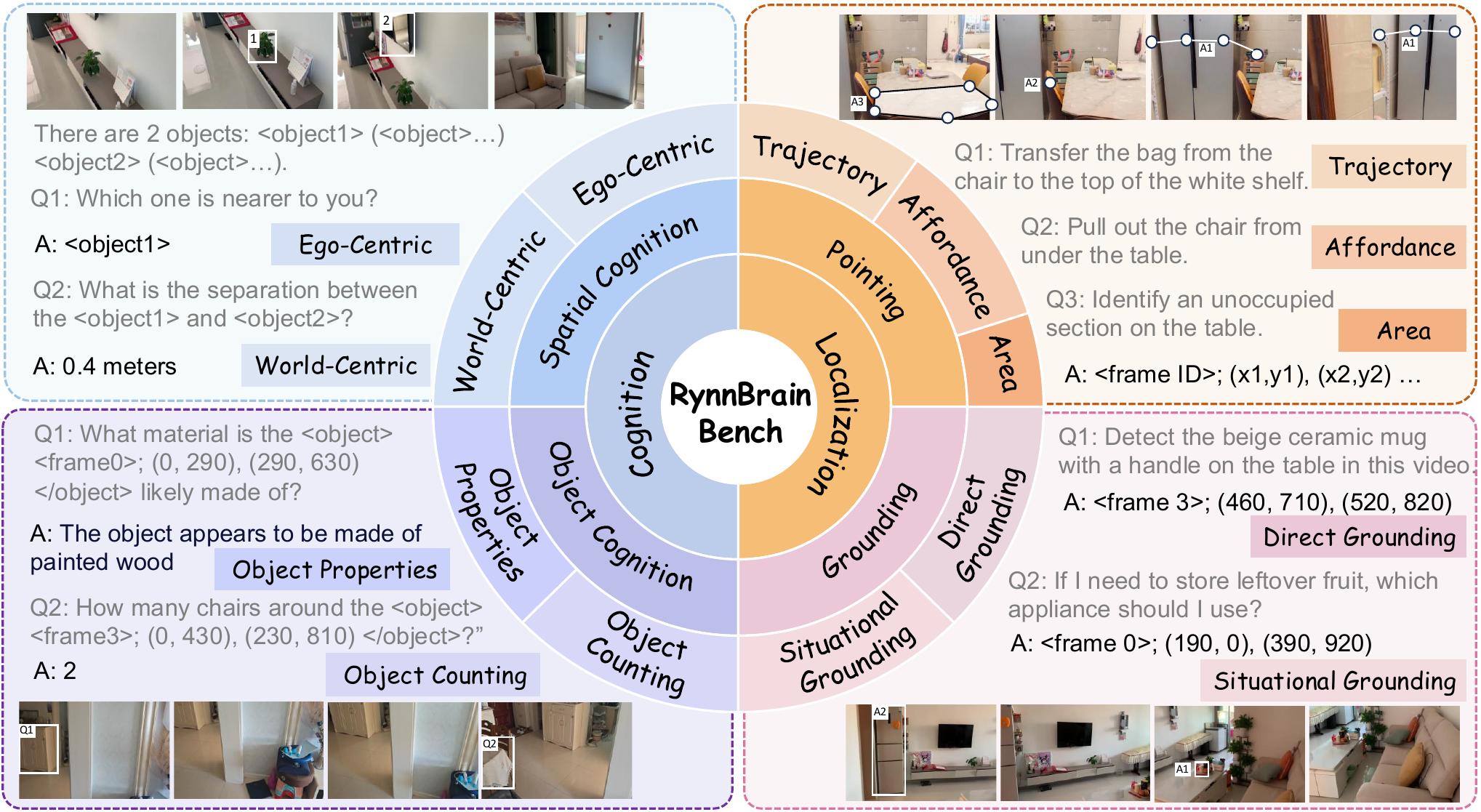}
    \caption{Overview of evaluation dimensions in RynnBrain-Bench. RynnBrain-Bench includes two subsets: cognition and location, evaluating a total of 21 spatio-temporal fine-grained embodied abilities.}
    \label{fig:rynnbrain_bench}
\end{figure*}

We present an overview framework of RynnBrain-Bench in \Cref{fig:rynnbrain_bench}, highlighting its core dimensions and sample tasks.
RynnBrain-Bench systematically measures spatio-temporal embodied understanding across four foundational pillars: \emph{Object Cognition}, \emph{Spatial Cognition}, \emph{Grounding}, and \emph{Pointing}.
Covering 21 specialized sub-capabilities ranging from detailed object attributes (e.g., color, shape) to affordance points prediction, the benchmark comprises 3,616 video clips consisting of 577,998 frames, and 12,000 meticulously curated open-ended questions for comprehensive evaluation. 
Our data construction starts from the self-collected egocentric indoor videos and object-centric Q/A pairs initially generated with foundation model priors, followed by the rigorous human-in-the-loop annotation pipeline detailed in Section~\ref{sec:pretraindata}.
The annotated data is cross-validated by human annotators across multiple rounds to ensure its correctness and high quality.
We perform internal data balancing across sub-capabilities within each foundational dimension to ensure fairness and objectivity for evaluation.
To ensure high fidelity, questions related to objects are further balanced against real-world object distributions for better authenticity.

\subsubsection{Evaluation Dimensions}
RynnBrain-Bench defines a new form of spatio-temporal evaluation paradigm, requiring models to perform instruction-guided cognition and localization anchored to precise spatial and temporal coordinates.

\paragraph{Object Cognition}
challenges models with fine-grained object perception and counting of region-level targets across dynamic image sequences. 
We assess nine core object attributes (i.e., category, color, material, shape, state, position, function, surface detail, and size)—plus a distinct object counting capability.
Models are required to provide responses conditioned on questions with precise spatio-temporal positions (i.e., frame index and spatial coordinates).
\emph{Evaluation Metrics:}
During evaluation, responses are scored by GPT-4o on a scale from 0 to 1, utilizing either a binary scheme or a multi-level system with 0.2-point increments.

\paragraph{Spatial Cognition}
requires models to derive 3D spatial awareness from egocentric video streams, spanning two primary perspectives: Ego-centric and World-centric. While ego-centric cognition examines the embodied agent's evolving relationship (e.g., rotation, direction) with the environment over time, world-centric cognition evaluates the comprehension of objective 3D layouts and physical properties, such as size scale, distance, and position. 
\emph{Evaluation Metrics:}
For numerical questions, we apply mean relative accuracy (MRA) and rotational accuracy (RoA) to measure the score following RynnEC~\cite{rynnec}.  
For textual questions, we use the binary or fine-grained scores from GPT-4o as described above.

\paragraph{Grounding} 
evaluates the capability for precise spatio-temporal localization, representing a key link for anchoring understanding in reality. This task requires the brain model to (1) pinpoint the critical temporal key frame and then (2) predict the object's spatial coordinates within that frame. We distinguish between Direct Grounding, which involves locating objects based on explicit descriptions, and Situational Grounding, which necessitates context-aware reasoning to identify and localize targets within complex scenarios.
\emph{Evaluation Metrics:}
We apply the Acc@0.5 to calculate the score~\cite{refcoco}. 
Specifically, the prediction is considered correct only if the model selects a frame $t$ that contains a valid ground truth ($\mathcal{G}_{t} \neq \emptyset$) and the Intersection over Union (IoU) between the predicted box $\mathcal{B}$ and $\mathcal{G}_{t}$ exceeds 0.5. 
Let $\mathbb{I}(\cdot)$ be the indicator function, the metric is:
\begin{equation}
\text{Acc@0.5} = \mathbb{I} \left( \mathcal{G}_{t} \neq \emptyset \land \text{IoU}(\mathcal{B}, \mathcal{G}_{t}) > 0.5 \right)
\end{equation}

\paragraph{Pointing} 
aims to predict target \emph{areas}, spatio-temporal \emph{trajectories}, or \emph{affordance} points across the entire episodic memory, serving as a critical bridge for robot-physical world interaction.
Departing from previous benchmarks, we extend the evaluation scope to the spatio-temporal domain, where models must demonstrate the dual capacity to locate the key frame and predict corresponding task-relevant point sequences.
\emph{Evaluation Metrics:}
For pointing tasks, the score is set to zero if the model-predicted frame does not contain a valid ground truth ($\mathcal{G}_{t} = \emptyset$). Otherwise, 
(1) For trajectory prediction, we apply the Discrete Fréchet Distance (DFD) distance~\cite{ji2025robobrain} between the predicted point sequence $\mathcal{P} = (p_1, \dots, p_M)$ and the ground truth sequence $\mathcal{G} = (g_1, \dots, g_N)$. 
We resample both sequences to 15 points uniformly distributed along the arc length and compute the DFD according to Equation~\ref{eq:dfd}.
(2) For area prediction, we calculate the proportion of predicted points $\mathcal{P}$ falling within the ground truth polygon $S_{\mathcal{G}}$ according to Equation~\ref{eq:area}.
(3) For affordance prediction, we evaluate spatial proximity using the exponential decay of the Euclidean distance from each predicted point $p \in \mathcal{P}$ to its nearest neighbor in the ground-truth set $\mathcal{G}$:
\begin{equation}
    D(\mathcal{P}, \mathcal{G}) = \operatorname{exp}(-\frac{1}{|\mathcal{P}|} \sum_{p \in \mathcal{P}} \min_{g \in \mathcal{G}} \|p - g\|_2)
\end{equation}

\begin{table}[t]
\caption{Comparison between models with parameter scales below 8B. * denotes results obtained from our own reproduction.}
\centering
\setlength{\tabcolsep}{4pt}
\renewcommand{\arraystretch}{1.5}

\begin{adjustbox}{width=\textwidth}
\begin{tabular}{
p{2.5cm} p{3.8cm}
!{\color{rynn}\vrule width 1.2pt}
>{\columncolor{rynn!7}\fontsize{11.5pt}{13.2pt}\selectfont\hspace{6pt}}c<{\hspace{8pt}} 
>{\columncolor{rynn!7}\fontsize{11.5pt}{13.2pt}\selectfont\hspace{8pt}}c<{\hspace{8pt}}
@{\color{rynn}\vrule width 1.2pt}
>{\hspace{8pt}}c<{\hspace{8pt}} 
c c c c
}
\toprule
\multicolumn{2}{l!{\color{rynn}\vrule width 1.2pt}}{%
    \multirow{2}{*}{%
        \diagbox[width=\dimexpr2.5cm+3.8cm+2\tabcolsep\relax, height=4em]{\textbf{Benchmark}}{\textbf{Model}}%
    }%
}
& \textbf{RynnBrain} & \textbf{RynnBrain}
& \makecell{\textbf{MiMo-}\\\textbf{Embodied}}
& \textbf{RoboBrain 2.0}
& \textbf{Pelican-VL}
& \textbf{Cosmos-reason2}
& \textbf{Qwen3-VL}
\\
& 
& \textbf{2B} & \textbf{8B}
& \textbf{7B} & \textbf{7B} & \textbf{7B} & \textbf{8B} & \textbf{8B}
\\
\midrule

\multirow{11}{*}{\makecell[l]{\textbf{Embodied}\\\textbf{Cognition}}}
& VSI-Bench                 & 70.5 & \textbf{71.0} & 48.5  & 36.1  & 52.8  & 53.7* & 60.3 \\
& MMSI                      & 34.1 & \textbf{39.6} & 30.2* & 24.8* & 26.2* & 31.3* & 29.6 \\
& ERQA                      & 42.3 & \textbf{46.8} & 46.8  & 36.5* & 39.8* & 46.0* & 44.8 \\
& RoboSpatial               & 65.7 & \textbf{73.1} & 61.8  & 54.2  & 57.5  & 59.0* & 58.2 \\
& EgoTaskQA                 & \textbf{73.9} & 72.5 & 58.7* & 51.1* & 50.0* & 55.7* & 57.8 \\
& \makecell[l]{EgoTextVQA$_{\text{indoor}}$}
                             & 27.7/2.08 & 31.6/2.28 & 28.7/2.17* & 22.0/1.79* & 30.3/2.24* & 26.5/1.96* & \textbf{38.9/2.64*} \\
& Open-X VQA                & 71.0 & \textbf{74.0} & 41.5* & 44.6  & 44.1* & 55.0* & 59.8 \\
& QAEgo4D                   & 43.9 & 43.9          & 39.0* & 39.7* & 26.1* & \textbf{46.9*} & 44.0 \\
& MindCube                  & 50.1 & \textbf{56.6} & 43.1* & 38.9  & 33.7* & 43.9* & 36.0 \\
& RynnBrain-Object          & 70.7 & \textbf{71.2} & 39.0  & 24.7  & 30.8  & 37.2  & 41.8 \\
& RynnBrain-Spatial         & 57.2 & \textbf{59.9} & 28.3  & 13.5  & 20.5  & 31.4  & 35.0 \\
\midrule

\multirow{9}{*}{\makecell[l]{\textbf{Embodied}\\\textbf{Location}}}
& RefSpatial-Bench          & 52.7 & \textbf{59.2} & 48.0  & 32.5  & 22.3  & 33.1* & 53.4 \\
& ShareRobot-Affordance     & 43.3 & \textbf{44.7} & 35.8* & 28.1  & 11.3  & 37.1* & 37.0 \\
& ShareRobot-Trajectory $\downarrow$
                             & \textbf{0.34} & 0.35 & 0.41* & 0.55  & 0.42* & 0.36* & 0.37 \\
& Cornell-Grasp             & 20.9 & \textbf{26.6} & 0.2*  & 0.0*  & 0.0*  & 18.1* & 21.2* \\
& VMRD-Grasp                & 13.0 & \textbf{14.1} & 2.8*  & 0.5*  & 0.0*  & 13.7* & 7.1* \\
& RynnBrain-Grounding       & 79.1 & \textbf{81.6} & 49.8  & 18.6  & 3.5   & 60.0  & 62.8 \\
& RynnBrain-Area            & 54.6 & \textbf{56.2} & 49.4  & 38.0  & 46.5  & 37.6  & 30.0 \\
& RynnBrain-Affordance      & 89.4 & \textbf{90.4} & 84.4  & 73.5  & 81.4  & 83.9  & 82.9 \\
& RynnBrain-Trajectory      & \textbf{66.6} & 64.5 & 61.3  & 57.6  & 59.2  & 64.0  & 63.4 \\
\midrule

\multirow{8}{*}{\makecell[l]{\textbf{General Visual}\\\textbf{Understanding}}}
& AI2D                      & 79.4 & \textbf{86.3} & 84.2  & 70.3* & 83.8* & 83.0* & 85.7 \\
& ChartQA                   & 78.2 & 86.5          & 85.2* & 82.4* & 87.5* & 84.3* & \textbf{89.6} \\
& DocVQA$_{\text{val}}$     & 93.0 & 96.2          & 94.9* & 93.1* & 94.5* & 95.0* & \textbf{96.4} \\
& MVBench                   & 67.3 & \textbf{69.5} & 57.9* & 50.6* & 67.7  & 67.0* & 68.7 \\
& RealWorldQA               & 60.4 & 67.3          & 66.1* & 51.2* & 67.1* & 69.3* & \textbf{71.5} \\
& InfoVQA$_{\text{test}}$   & 71.2 & \textbf{83.4} & 72.0* & 77.6* & 81.1* & 78.3* & 83.1 \\
& EgoSchema                 & 64.0 & 69.7          & 58.2* & 54.2* & \textbf{73.3} & 63.5* & 69.7 \\
& VideoMME$_{\text{w/o sub}}$
                             & 61.4 & 70.7          & 65.0* & 52.3* & 63.3  & \textbf{71.9} & 71.4 \\
\bottomrule
\end{tabular}
\end{adjustbox}

\label{tab:rynnbrain-8b}
\end{table}

\begin{table}[t]
\caption{Comparison between models with parameter scales above 30B. * denotes results obtained from our own reproduction.}
\centering
\setlength{\tabcolsep}{4pt}
\renewcommand{\arraystretch}{1.5}

\begin{adjustbox}{width=\textwidth}
\begin{tabular}{
p{2.5cm} p{3.8cm}
!{\color{rynn}\vrule width 1.2pt}
>{\columncolor{rynn!7}\fontsize{11.5pt}{13.2pt}\selectfont}c<{\hspace{4pt}} 
@{\color{rynn}\vrule width 1.2pt}
>{\hspace{4pt}}c<{\hspace{6pt}} 
c c c c c
}
\toprule
\multicolumn{2}{l!{\color{rynn}\vrule width 1.2pt}}{%
    \multirow{2}{*}{%
        \diagbox[width=\dimexpr2.5cm+4cm+2\tabcolsep\relax, height=3.8em]{\textbf{Benchmark}}{\textbf{Model}}%
    }%
}
& \textbf{\large RynnBrain} 
& \textbf{RoboBrain 2.0} 
& \textbf{Pelican-VL} 
& \textbf{GPT-5.2} 
& \textbf{Gemini 3 Pro} 
& \textbf{Claude Sonnet 4.5} 
& \textbf{Qwen3-VL} 
\\
& 
& \textbf{30B (A3B)} 
& \textbf{32B} 
& \textbf{72B} 
& \textbf{-} 
& \textbf{-} 
& \textbf{-} 
& \textbf{30B (A3B)} 
\\
\midrule

\multirow{11}{=}{\textbf{Embodied\\Cognition}} 
& VSI-Bench              & \textbf{74.5} & 42.7 & 57.3 & 46.6* & 48.8* & 42.5* & 65.8 \\
& MMSI                   & 39.5 & 28.5* & 30.7* & 38.2* & \textbf{49.2} & 28.9* & 21.1 \\
& ERQA                   & 46.3 & 46.0 & 43.0 & 45.3* & \textbf{70.5} & 60.0 & 43.0 \\
& RoboSpatial            & 70.0 & \textbf{72.4} & 55.4 & 54.7* & 56.0* & 40.9* & 55.4 \\
& EgoTaskQA              & \textbf{78.9} & 59.9* & 64.8* & 59.6* & 68.4* & 50.9* & 64.2* \\
& EgoTextVQA\textsubscript{indoor} & 34.6/2.39 & 30.5/2.24* & 37.8/2.59* & \textbf{49.6/3.02}* & 45.5/2.87* & 36.6/2.55* & 41.3/2.74* \\
& Open-X VQA             & \textbf{83.4} & 28.6* & 48.0* & 43.6* & 56.0* & 41.9* & 76.8* \\
& QAEgo4D                & \textbf{47.3} & 40.3* & 24.6* & 46.8* & 42.1* & 35.0* & 47.3* \\
& MindCube               & 63.4 & 29.2 & 32.5* & 61.7 & \textbf{70.8} & 58.3 & 39.0* \\
& RynnBrain-Object       & \textbf{73.3} & 26.2 & 42.2 & 53.1 & 44.6 & 25.1 & 42.6 \\
& RynnBrain-Spatial      & \textbf{59.3} & 11.6 & 32.2 & 33.7 & 29.0 & 34.2 & 30.7 \\
\midrule

\multirow{9}{=}{\textbf{Embodied\\Location}} 
& RefSpatial-Bench       & 59.2 & 54.0 & 49.5 & 26.4* & \textbf{65.5} & 15.1 & 53.1 \\
& ShareRobot-Affordance  & 43.2 & 35.3 & 10.4* & 17.5* & 26.9* & 13.9 & \textbf{47.2}* \\
& ShareRobot-Trajectory $\downarrow$ & 0.31 & \textbf{0.24} & 0.36* & 0.35* & 0.29* & 0.57 & 0.36* \\
& Cornell-Grasp          & \textbf{33.6} & 0.3* & 0.0* & 14.5* & 33.2* & 0.0* & 29.9* \\
& VMRD-Grasp             & \textbf{14.5} & 0.7* & 0.0* & 6.2* & 10.9* & 4.8* & 8.0* \\
& RynnBrain-Grounding    & \textbf{83.9} & 0.0 & 10.8 & 11.2 & 59.2 & 0.0 & 76.4 \\
& RynnBrain-Area         & 59.4 & 45.3 & 53.2 & 35.8 & \textbf{61.5} & 10.1 & 30.9 \\
& RynnBrain-Affordance   & \textbf{90.5} & 76.1 & 87.3 & 83.3 & 86.0 & 48.7 & 86.2 \\
& RynnBrain-Trajectory   & 66.8 & 60.3 & 64.1 & 70.5 & \textbf{72.0} & 54.6 & 61.2 \\
\midrule

\multirow{7}{=}{\textbf{General\\Visual\\Understand}} 
& AI2D                   & 87.0 & 67.3* & 86.7 & 97.1 & \textbf{98.7} & 91.5 & 85.0 \\
& ChartQA                & 88.3 & 82.4* & 90.4* & 89.6 & \textbf{93.7} & 88.1 & 83.7* \\
& DocVQA                 & \textbf{96.3} & 90.2* & 95.2* & 94.2 & 87.1 & 91.7 & 95.0 \\
& MVBench                & 70.8 & 57.1* & 69.7 & 67.1* & 71.5* & 55.1* & \textbf{72.3} \\
& RealworldQA            & 69.7 & 67.5* & 67.3* & \textbf{82.5}* & 73.6* & 68.1 & 73.7 \\
& InfoVQA\textsubscript{test} & 83.1 & 75.5* & \textbf{89.1}* & 66.8* & 83.1* & 62.2* & 82.0 \\
& EgoSchema              & 66.8 & 61.3* & 79.3 & \textbf{81.2}* & 72.2* & 67.2* & 70.7 \\
& VideoMME\textsubscript{w/o sub} & 71.9 & 55.6* & 73.7* & 84.7 & \textbf{88.6} & 68.6* & 74.5 \\

\bottomrule
\end{tabular}
\end{adjustbox}

\label{tab:rynnbrain-30b}
\end{table}







\subsection{Embodied Cognition Capability}
\todo[inline, color=myred!20]{Assigned to: Yuqian Yuan}
To assess RynnBrain’s embodied cognition capabilities, we evaluate it on a diverse suite of benchmarks, including VSI-Bench~\cite{yang2025thinking}, MMSI~\cite{yang2025mmsi}, ERQA~\cite{team2025gemini}, RoboSpatial~\cite{song2025robospatial}, EgoTaskQA~\cite{jia2022egotaskqa}, EgoTextVQA~\cite{zhou2025egotextvqa}, Open-X VQA~\cite{chen2025robo2vlm}, MindCube~\cite{yin2025spatial}, RynnBrain-Object and RynnBrain-Spatial. As shown in \Cref{tab:rynnbrain-8b}, our RynnBrain-8B outperform the base model Qwen3-VL-8B on 9 of 11 embodied cognition tasks. RynnBrain-8B delivers substantial gains across a variety of tasks. For instance, on the spatial reasoning benchmark VSI-Bench, RynnBrain-8B achieves 71.0 score, surpassing the previous best result of 60.3, and on RoboSpatial it exceeds the previous top-performing method by 11.3\%. RynnBrain-8B also attains strong performance on RynnBrain-Object and RynnBrain-Spatial, indicating robust improvements in object-centric and spatially grounded reasoning. 
Similarly, we evaluate RynnBrain-30B (A3B) model on various embodied cognition benchmarks, with results summarized in \Cref{tab:rynnbrain-30b}. From the table, it is clear that RynnBrain-30B (A3B) outperforms prior models on most benchmarks. Notably, on VSI-Bench it improves over the previous best by 8.7\%, on EgoTaskQA it yields a 10.5\% gain, on Open-X VQA it surpasses prior methods by 6.6\%, on RynnBrain-Object it improves by 20.2\%, and on RynnBrain-Spatial it achieves a 25.1\% gain. These results collectively demonstrate RynnBrain’s strong ability to perform embodied cognition and spatial reasoning across diverse tasks and environments.

\subsection{Embodied Location Capability}
\todo[inline, color=mygreen!20]{Assigned to: Zhikai Wang}
We evaluate RynnBrain's spatial grounding abilities across five key location tasks: object location, area location, affordance location,  trajectory location, and grasp pose location. Our models are benchmarked against state-of-the-art methods on public benchmarks including RefSpatial-Bench~\cite{zhou2025roborefer}, ShareRobot-Affordance~\cite{ji2025robobrain}, ShareRobot-Trajectory~\cite{ji2025robobrain}, Cornell-Grasp~\cite{chu2018deep}, and VMRD-Grasp~\cite{zhang2019roi}.
As shown in \Cref{tab:rynnbrain-8b}, RynnBrain-8B achieves leading performance across all location benchmarks except ShareRobot-Trajectory, where RynnBrain-2B performs best. On RefSpatial-Bench, it achieves 59.2, surpassing the base model~(Qwen3-VL) by 5.8\%. It attains 44.7 on ShareRobot-Affordance, outperforming the closest competitor by 7.7\%. For grasp pose location, RynnBrain-8B achieves 26.6 on Cornell-Grasp and 14.1 on VMRD-Grasp, significantly exceeding other 8B-scale models.  Moreover, on our internal RynnBrain-Grounding and RynnBrain-Affordance benchmarks, RynnBrain-8B reaches 81.6 and 90.4, respectively, demonstrating its strong capability in precise spatio-temporal joint localization.
The advantages are kept at the 30B (A3B) scale (~\Cref{tab:rynnbrain-30b}). It achieves the best results on Cornell-Grasp (33.6), VMRD-Grasp (14.5), RynnBrain-Grounding (83.9), and RynnBrain-Affordance (90.5). 
Moreover, RynnBrain-30B (A3B) significantly outperforms all other models of the same scale on RefSpatial-Bench, RynnBrain-Area, and RynnBrain-Trajectory, approaching the performance of the much larger Gemini 3 Pro.
These results demonstrate RynnBrain's strong spatial grounding capabilities across multiple embodied location tasks and model scales.

\subsection{General Visual Understanding}
\todo[inline, color=myred!20]{Assigned to: Yuqian Yuan}
We further evaluate the general visual understanding ability of RynnBrain to assess its overall generality and generalization. To cover both static images and dynamic videos, we benchmark RynnBrain on a suite of general VQA datasets, including image-based AI2D~\cite{kembhavi2016diagram}, ChartQA~\cite{masry2022chartqa}, DocVQA~\cite{mathew2021docvqa}, RealWorldQA~\cite{realworldqa}, and InfoVQA~\cite{mathew2022infographicvqa}, as well as video-based MVBench~\cite{li2024mvbench}, EgoSchema~\cite{mangalam2023egoschema}, and VideoMME~\cite{fu2025video}. As shown in \Cref{tab:rynnbrain-8b}, RynnBrain maintains the general visual understanding performance of the base model Qwen3-VL on both images and videos, and notably achieves state-of-the-art results on AI2D, MVBench, and InfoVQA, demonstrating the effectiveness of our training strategy. \Cref{tab:rynnbrain-30b} shows consistent trends for the 30B models, confirming that RynnBrain provides strong general visual capability alongside its embodied cognition strengths. This generalization advantage enables RynnBrain to serve as a central component of an embodied agent system, accommodating diverse task requirements.

\subsection{Physically Grounded Reasoning}
\todo[inline, color=myblue]{Assigned to: Jiayan Guo}

\begin{table}[t!]
    \centering
    \caption{Comparison RynnBrain-CoP with state-of-the-art thinking models on embodied reasoning tasks. All compared models are evaluated with the thinking mode enabled. We evaluate models on affordance prediction, area prediction, and trajectory prediction. Average denotes the mean across the three tasks. Best results are highlighted.}
    \resizebox{0.85\linewidth}{!}{
    \begin{tabular}{l|cccc}
    \toprule
    \multirow{2}{*}{Model} & \multicolumn{4}{c}{Task} \\
    & Affordance & Area & Trajectory & Average \\
    \midrule
    InternVL3.5-8B~\cite{internvl3_5}          & 63.1 &  9.2 & 47.8 & 40.0 \\
    MiMo-Embodied-7B~\cite{MiMo-Embodied}        & 85.3 & 47.1 & 64.9 & 65.8 \\
    RoboBrain2.0-7B~\cite{robobrain2.0}         & 65.3 & 38.0 & 58.5 & 53.9 \\
    RoboBrain2.0-32B~\cite{robobrain2.0}        & 73.2 & 39.5 & 60.5 & 57.7 \\
    Qwen3-VL-8B-Thinking~\cite{qwen3vl}    & 56.7 & 20.4 & 46.9 & 41.3 \\
    Qwen3-VL-30B-A3B-Thinking~\cite{qwen3vl} & 62.2 & 33.0 & 54.8 & 50.0 \\
    GPT-5.2~\cite{singh2025openai}                 & 83.3 & 35.8 & 70.5 & 63.2 \\
    Gemini-3-Pro            & 83.9 & 50.7 & 60.6 & 65.1 \\
    \arrayrulecolor{rynn}\midrule[1pt]\arrayrulecolor{black}
\rowcolor{rynn!10} 
\textbf{\fontsize{11.5}{13.6}\selectfont RynnBrain-CoP-8B}  & \textbf{90.3} & \textbf{59.6} & \textbf{71.2} & \textbf{73.8}   \\
    \bottomrule
    \end{tabular}
    }
    \label{tab:comparison_of_reasoning}
\end{table}


To rigorously evaluate the physically grounded reasoning capabilities of our model, we conducted a comparative analysis of RynnBrain-CoP-8B against several state-of-the-art multimodal baselines. The comparison includes leading open-source models such as InternVL3.5-8B~\cite{internvl3_5}, MiMo-Embodied-7B~\cite{MiMo-Embodied}, and Qwen3-VL~\cite{qwen3vl} (8B and 30B variants), alongside powerful proprietary models like GPT-5.2 and Gemini-3-Pro. Our evaluation focuses on three core embodied tasks—affordance prediction, area prediction, and trajectory prediction—which require the model to ground complex spatial intent into precise coordinates.

As shown in ~\Cref{tab:comparison_of_reasoning}, RynnBrain-CoP-8B achieves superior performance across all evaluated metrics, setting a new state-of-the-art for embodied reasoning. On average, our 8B model attains a score of 73.8, surpassing the strongest proprietary competitor, MiMo-Embodied-7B (65.8) and Gemini-3-Pro (65.1), by a substantial margin. Notably, it outperforms the much larger RoboBrain2.0-32B (57.7) by 16.1\%, demonstrating that our reasoning architecture is more effective than simple parameter scaling for spatial tasks.

The task-specific results further highlight the model's precision:
\begin{itemize}
    \item Affordance Prediction: RynnBrain-CoP-8B achieves a peak accuracy of 90.3, being the only model to break the 90 threshold. This suggests that the physically grounded CoT effectively narrows down actionable zones.
    \item Area Prediction: While this remains the most challenging task for all baselines (with many scoring below 40), our model reaches 59.6, outperforming Gemini-3-Pro (50.7) and nearly doubling the performance of Qwen3-VL-30B (33.0).
    \item Trajectory Prediction: Our model leads with 71.2, showcasing a superior understanding of temporal-spatial sequences compared to GPT-5.2 (70.5) and InternVL3.5 (47.8).
\end{itemize}

These results validate that despite its compact 8B parameter size, RynnBrain-CoP-8B delivers consistently more accurate spatial grounding. The significant gains, particularly in complex area and trajectory tasks, prove that interleaving multi-step reasoning thoughts with visual coordinates is a more data-efficient and hallucination-resistant paradigm for embodied agents than traditional purely text-based reasoning paradigm.

\subsection{Vision-Language Navigation}
\todo[inline, color=mypurple!20]{Assigned to: Jiangpin Liu. 
The comparison model: NaVILA, uniNavid, StreamVLN, Gemini 3 Pro, GPT 5.2, Qwen3VL}

\begin{table*}[t]
\caption{Comparison RynnBrain-Nav with state-of-the-art navigation models. The best results are highlighted.}
\centering
\resizebox{\textwidth}{!}{%
\begin{tabular}{l | cccc | cccc | cccc}
\toprule

\multirow{2}{*}{Method} & \multicolumn{4}{c|}{Observation Encoder} & \multicolumn{4}{c|}{R2R Val-Unseen} & \multicolumn{4}{c}{RxR Val-Unseen} \\
\cmidrule(lr){2-5} \cmidrule(lr){6-9} \cmidrule(l){10-13}
 & Pano. & Odo. & Depth & S.RGB & NE$\downarrow$ & OS$\uparrow$ & SR$\uparrow$ & SPL$\uparrow$ & NE$\downarrow$ & SR$\uparrow$ & SPL$\uparrow$ & nDTW$\uparrow$ \\
\midrule

VLN$\circlearrowright$BERT*~\cite{VLN_BERT*} & \checkmark & \checkmark & \checkmark & & 5.74 & 53.0 & 44.0 & 39.0 & 8.98 & 27.0 & 22.6 & 46.7 \\
ETPNav*~\cite{etpnav} & \checkmark & \checkmark & \checkmark & & 4.71 & 65.0 & 57.0 & 49.0 & 5.64 & 54.7 & 44.8 & 61.9 \\
ScaleVLN*~\cite{scalevln} & \checkmark & \checkmark & \checkmark & & 4.80 & -- & 55.0 & 51.0 & - & - & - & - \\

\hline 

R2R-CMTP~\cite{R2R-CMTP} &\checkmark & \checkmark & \checkmark & & 7.90 & 38.0 & 26.4 & 22.7 & - & - & - & - \\

LAW~\cite{law} & & \checkmark & \checkmark & \checkmark & 6.83 & 44.0 & 35.0 & 31.0 & 10.90 & 8.0 & 8.0 & 38.0 \\
ETPNav + FF~\cite{etpnav+ff} & & \checkmark & \checkmark & \checkmark & 5.95 & 55.8 & 44.9 & 30.4 & 8.79 & 25.5 & 18.1 & - \\
Seq2Seq~\cite{Seq2Seq} & & & \checkmark & \checkmark & 7.77 & 37.0 & 25.0 & 22.0 & 12.10 & 13.9 & 11.9 & 30.8 \\
CMA~\cite{Seq2Seq} & & & \checkmark & \checkmark & 7.37 & 40.0 & 32.0 & 30.0 & - & - & - & - \\
\hline
VLN-R1~\cite{VLNR1} & & &  & \checkmark & 5.47 & 49.1 & 37.4 & 35.9 & - & - & - & - \\

NaVid~\cite{NaVILA} & & &  & \checkmark & 5.47 & 49.1 & 37.4 & 35.9 & - & - & - & - \\


NaVILA~\cite{NaVILA}& & &  & \checkmark & 5.22 & 62.5 & 54.0 & 49.0 & 6.77 & 49.3 & 44.0 & 58.5 \\
UniNaVid~\cite{UniNavid} & & &  & \checkmark &5.58 &53.3 &47.0& 42.7 &6.24 &48.7 &40.9&- \\
StreamVLN~\cite{wei2025streamvln} & & &  & \checkmark & 4.98 & 64.2 & 56.9 & \textbf{51.9} & 6.22& 52.9 & 46.0 & \textbf{61.9} \\
\arrayrulecolor{rynn}\midrule[1pt]\arrayrulecolor{black}
\rowcolor{rynn!10} 
\textbf{\fontsize{11.5}{13.6}\selectfont RynnBrain-Nav-8B} & & & &\checkmark &\textbf{4.92} & \textbf{71.6}& \textbf{58.6}& 49.6 &\textbf{6.20 }&\textbf{56.1} & \textbf{49.6}&59.6 \\

\bottomrule
\end{tabular}
}
\label{tab:vln_ce}
\end{table*}


\textbf{Benchmarks and Metrics.} We evaluate our finetuned model on two public VLN-CE benchmarks~\cite{Seq2Seq}: R2R-CE~\cite{r2r-ce} and RxR-CE~\cite{rxr-ce}, which simulate continuous navigation in photorealistic Matterport3D scenes using the Habitat simulator. To assess generalization to novel environments, all experiments are conducted on the validation unseen splits. Following standard protocols, we report performance using metrics for task completion (Success Rate, SR), path efficiency (Success-weighted by Path Length, SPL), and path fidelity (normalized Dynamic Time Warping, nDTW). The nDTW metric, specifically, leverages the ground-truth trajectories to evaluate how closely the agent's path follows the reference instruction. We also include Navigation Error (NE) and Oracle Success Rate (OS) for a comprehensive analysis.

\textbf{Comparision wth State-of-The-Art.} \Cref{tab:vln_ce} summarizes the performance of our method on the VLN-CE R2R and RxR benchmarks under the Val-Unseen setting, compared with existing SOTA methods.
For the R2R-CE benchmark, the RynnBrain-Nav-8B model demonstrates highly competitive performance even compared to methods utilizing multiple input types like panoramic views and odometry. Achieving a top-ranked SR of 58.6\% and the second-best SPL of 49.6\% and the lowest NE of 4.92.  Noticing that our model's OS reaching 71.6\% exceeds all competitors, including topological prediction methods that utilize panoramic observations. This contrast between our high OS and lower SR indicates that our model is proficient at coarse-level navigation but lacks the precision for the terminal stopping maneuver, thereby failing the overall task. 

The model's navigation capability is further validated on the more demanding RxR benchmark. Here, RynnBrain-Nav-8B again secures a top-ranked SR of 56.1\% and the lowest NE of 4.92, highlighting its superior capability in complex, long-horizon navigation tasks.

\begin{figure}[!t]
    \centering
    \includegraphics[width=\linewidth]{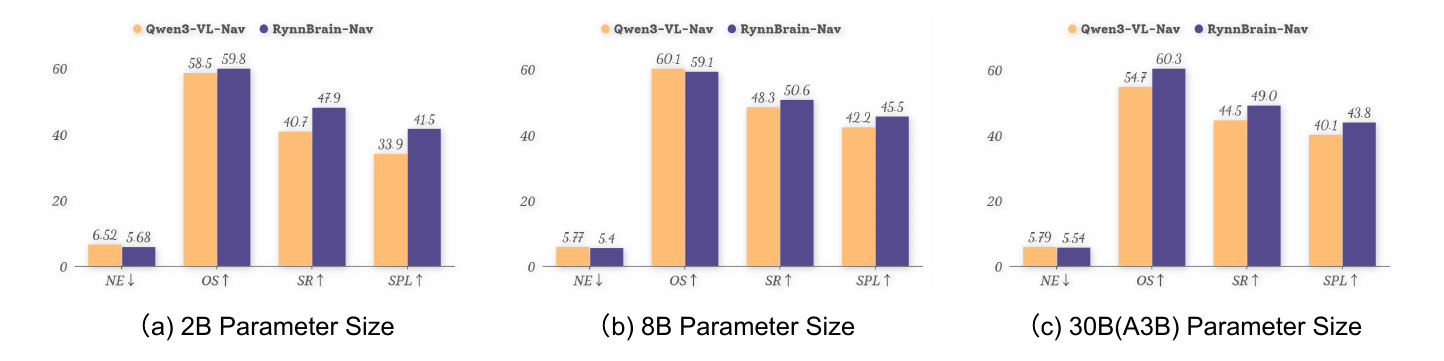}
    \caption{Compare the differences in the ability of Qwen3-VL and RynnBrain as the base model to finetune navigation models under multiple model scales. All results are reported without performing multiple rounds of DAgger.}
    \label{fig:nav_compare}
\end{figure}

\textbf{Ablation Study on Pre-training Efficacy.} To isolate and evaluate the contribution of our pretraining, we conduct a comparative analysis between RynnBrain and its Qwen3-VL~\cite{qwen3vl} baseline. Models of varying scales from both families are fine-tuned on the same sample of datasets (R2R, RxR, EnvDrop, ScaleVLN), and their performance is evaluated  on the R2R-CE benchmark. The results, shown in \Cref{fig:nav_compare}, demonstrate the benefit of the RynnBrain pretraining. RynnBrain-Nav demonstrates clear performance superiority over the Qwen3-VL counterpart, achieving consistently higher SR and SPL scores across all evaluated scales. Notably, our 2B RynnBrain-Nav model surpasses its 2B Qwen3-VL counterpart by a substantial 7.2\% in SR and 7.6\% in SPL, affirming the clear efficacy of our pretraining approach.

\textbf{Impact of Model Scale and Architecture.} Our analysis reveals a clear scaling trend for dense architectures. As shown in \Cref{fig:nav_compare}, both RynnBrain-Nav and Qwen3-VL demonstrate improved SR and SPL when scaling from 2B to 8B parameters. However, this positive scaling did not extend to the Mixture-of-Experts (MoE) architecture. Despite its larger total parameter count, the 30B MoE model (3B active) failed to outperform the 8B dense models during initial training phases. This suggests that the sparse activation mechanism of MoE may not be fully leveraged by the Visual Language Navigation (VLN) task, or that alternative training strategies are required to unlock its scaling potential.

 \textbf{Multi-Turn DAgger Training.}
To further enhance navigation performance, we employ multi-turn DAgger~\cite{dagger} training. After initial SFT, the agent collects new trajectories from the R2R, RxR, and EnvDrop environments. This data is then combined with the original datasets to retrain the model. This iterative process proved highly effective, particularly in the initial rounds: the Success Rate (SR) increased from a 50.6\% baseline to 56.4\% after the first iteration and further to 58.5\% after the second. However, the third DAgger iteration yielded only a marginal improvement, indicating a clear trend of diminishing returns as the agent's policy converges.

\subsection{Planning and Manipulation}
We develop a three-stage evaluation system to rigorously assess the hierarchical manipulation system based on RynnBrain.
In the first setting, we evaluate the planning logic in isolation: our model serves as the high-level planner, while a human operator equipped with a Universal Manipulation Interface (UMI)~\cite{chi2024universal} acts as a fully reliable low-level controller. 
In the second setting, we design three real-robot experiments in complex, multi-objective scenarios to validate the precise manipulation capabilities of RynnBrain-VLA.
In the third setting, we assess end-to-end autonomy by deploying the integrated system on a Franka robot. Throughout the entire system, \textbf{RynnBrain-Plan} is responsible for comprehending the scene and high-level tasks and then generating sub-tasks with precise coordinates. \textbf{RynnBrain-VLA} accepts sub-tasks and controls the robot arm to perform the low-level tasks.

\subsubsection{RynnBrain and UMI Hierarchical Evaluation}
\todo[inline, color=mybrown!20]{Assigned to: Yunxuan Mao}

\noindent \textbf{Experimental settings.} We designed four long-horizon planning tasks: Object Classification, Desk Organization, Distribute Tableware, and Table Bussing. Among them, the first three are in-distribution tasks, whereas the last one is an out-of-distribution task. To assess performance across varying degrees of complexity, all tasks are stratified into three difficulty levels: Easy, Medium, and Hard. As the difficulty level increases, both the scene complexity and the instruction complexity rise accordingly. For the fine-tuning phase, we collected 100 expert demonstrations for each of the in-distribution tasks. Detailed descriptions and specifications for each task are provided in the Appendix. We benchmark our method, RynnBrain-Plan, against two state-of-the-art baselines: Gemini-3 Pro, and Qwen3-VL 30B. To mitigate randomness, each task–model evaluation is repeated five times and we report the average results. Following the protocol established in~\cite{shi2025hirobotopenendedinstruction}, we adopt Task Progress—defined as the percentage of subtasks successfully completed by the end of the episode—as our primary evaluation metric. To guarantee reliability and consistency, all assessments are conducted by trained human annotators.

\begin{figure}
    \centering
    \includegraphics[width=\linewidth]{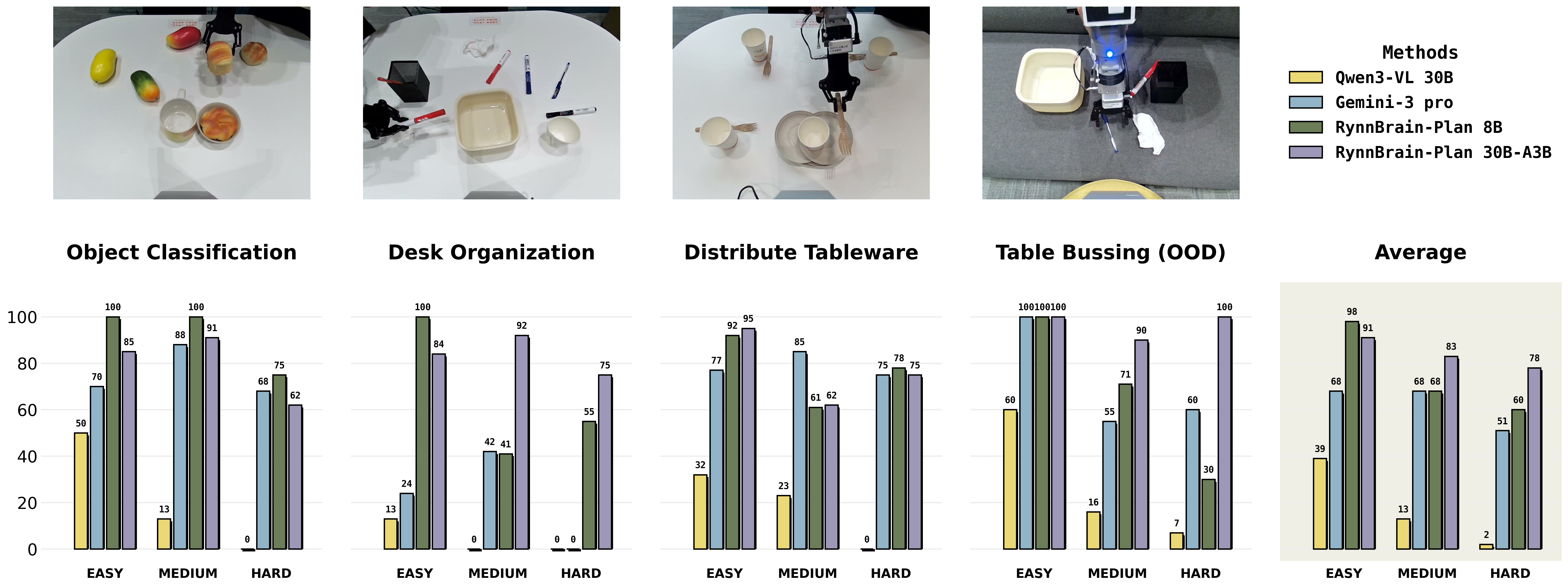}
    \caption{Comparison to other VLMs. RynnBrain-Plan-30B outperforms other methods on almost all the settings, except for the medium difficulty of Distribute Tableware. The metric is Task Process (TP $\uparrow$).}
    \label{fig:planning_comparison}
\end{figure}

\noindent \textbf{Comparison on ID Tasks.} As quantitatively illustrated in \Cref{fig:planning_comparison}, our Rynnbrain-Plan demonstrates a significant performance advantage over state-of-the-art baselines across varying difficulty levels. In the in-distribution tasks (\textit{Object Classification}, \textit{Desk Organization}, and \textit{Distribute Tableware}), our Rynnbrain-Plan-30B-A3B consistently achieves superior task progress. This advantage is particularly pronounced in the ``Hard'' difficulty settings, which require complex long-horizon reasoning. For instance, in the \textit{Desk Organization} task (Hard), while Qwen3-VL and Gemini-3 Pro fail to make meaningful progress (near 0\% completion), our 30B model maintains a robust completion rate of over 75\%. While Gemini 3 Pro shows competitive performance in simpler scenarios (e.g., \textit{Distribute Tableware} - Medium), it suffers from severe performance degradation as task complexity increases. The Rynnbrain-Plan 8B model also delivers strong results in ``Easy'' and ``Medium'' settings, often surpassing the significantly larger Qwen3-VL 30B, highlighting the efficiency of our data construction strategy. 

\noindent \textbf{Generalization Analysis on OOD Tasks.} The results on the out-of-distribution (OOD) task, \textit{Table Bussing}, highlight the exceptional generalization capabilities of our approach. Despite not being exposed to this specific task during fine-tuning, Rynnbrain-Plan 30B (A3B) achieves remarkable success, reaching near 100\% task progress across all difficulty levels. This stands in stark contrast to the baselines; for example, in the ``Hard'' setting of \textit{Table Bussing}, Qwen3-VL completely fails ($<10\%$), and Gemini-3 Pro achieves only moderate success ($\sim60\%$). Crucially, comparing our two model variants reveals that while the 8B model generalizes well in simple OOD scenarios, the larger 30B model possesses the emergent capacity to handle complex, unseen constraints, effectively bridging the gap between in-domain planning and open-world adaptability.

\noindent \textbf{Ablation Study.} To rigorously validate the effectiveness of fine-tuning with multi-turn dialogue data, we conducted an ablation experiment by training a variant of RynnBrain-Plan exclusively on single-turn dialogue samples. As presented in the~\Cref{tab:ablate_plan}, the performance of this single-turn baseline degrades significantly. It only manages to complete tasks in the ``Easy'' difficulty setting, yet even in these simple scenarios, the success rate remains prohibitively low. This sharp decline underscores the necessity of temporal context: without the multi-turn interaction history, the model struggles to maintain state consistency over time. In contrast, the model fine-tuned on multi-turn data effectively leverages historical actions to ground its reasoning, leading to substantially more accurate and coherent action predictions.

\begin{table*}[t]
\caption{Ablation on multi-turn dialogue data. Training with single-turn dialogue and multi-turn dialogue data is short for ST and MT. The metric is Task Process (TP $\uparrow$).}
\centering
\resizebox{\textwidth}{!}{ 
\begin{tabular}{l | ccc| ccc | ccc | ccc}
\toprule

\multirow{2}{*}{Method} & \multicolumn{3}{c|}{Object Classification} & \multicolumn{3}{c|}{Desk Organization} & \multicolumn{3}{c|}{Distribute Tableware} & \multicolumn{3}{c}{Table Bussing} \\
\cmidrule(lr){2-4} \cmidrule(lr){5-7} \cmidrule(l){8-10} \cmidrule(l){11-13} 
 & Easy & Medium & Hard & Easy & Medium & Hard & Easy & Medium & Hard & Easy & Medium & Hard \\
\midrule

RynnBrain-Plan-ST 8B & 72 & 20 & 0 & 60 & 0 & 0 & 34 & 0 & 0 & 90 & 0 & 0 \\
RynnBrain-Plan-ST 30B & 75 & 30 & 0 & 58 & 10 & 0 & 28 & 0 & 0 & 95 & 0 & 0 \\

\arrayrulecolor{rynn}\midrule[1pt]\arrayrulecolor{black}
\rowcolor{rynn!25} 
\rowcolor{rynn!10} \textbf{\fontsize{11.5}{13.6}\selectfont RynnBrain-Plan-MT 8B} & 100 & 100 & 75 & 100 & 41 & 55 & 92 & 61 & 78 & 100 & 71 & 30 \\
\rowcolor{rynn!10} \textbf{\fontsize{11.5}{13.6}\selectfont RynnBrain-Plan-MT 30B} & 85 & 91 & 62 & 84 & 92 & 75 & 95 & 62 & 75 & 100 & 90 & 100 \\

\bottomrule
\end{tabular}
}

\label{tab:ablate_plan}
\end{table*}




\begin{figure}
    \centering
    \includegraphics[width=1.0\linewidth]{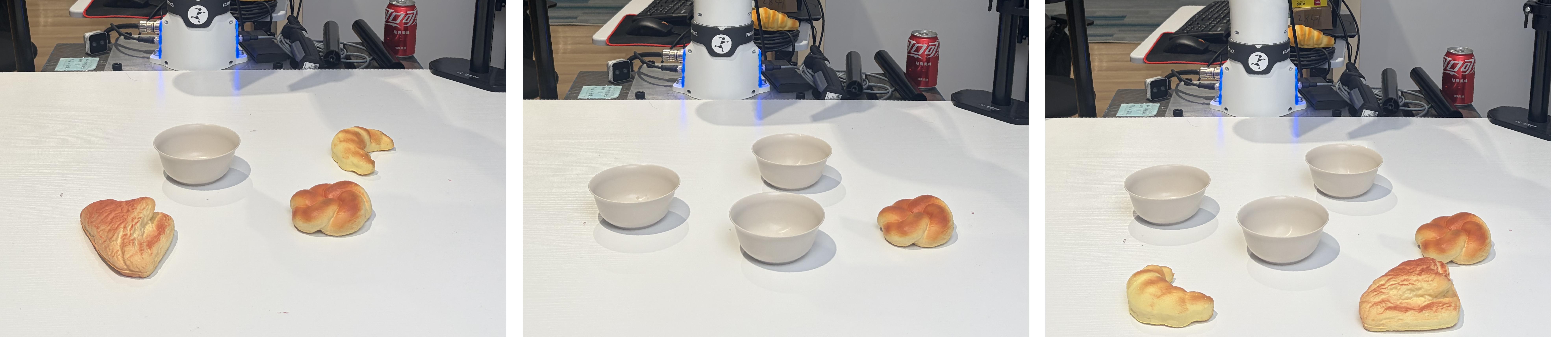}
    \caption{Tasks for VLA evaluation.}
    \label{fig:vla_eval}
\end{figure}

\subsubsection{VLA Evaluation.}
\todo[inline, color=mygray!20]{Assigned to: Kehan Li}

We benchmark RynnBrain-VLA in three multi-object scenarios to evaluate its object manipulation capabilities.
As illustrated in \Cref{fig:vla_eval}, the experimental setup includes two fundamental tasks featuring four manipulable objects from two categories (with one category dominant), and a more challenging task involving six objects with a balanced category distribution. Each model was tested over ten trials per task, with object arrangements and target selections randomized for each run.
To provide a comprehensive analysis, we employed three evaluation metrics:
(1) Pickup Success Rate (PSR): The percentage of trials where any object was successfully grasped, regardless of its identity.
(2) Recognition Success Rate (RSR): The accuracy in identifying the target object, defined by whether the gripper makes initial contact with the correct item.
(3) Success Rate (SR): The overall rate of successfully picking up the designated target object.
For comparative analysis, we fine-tuned two baseline models:
(1) $\pi$-0.5~\cite{pi_0.5}: In order to enable it to manipulate specific objects, we adapted its input by appending the initial task frame and employing a consistent text format as RynnBrain-VLA.
(2) Qwen3-VL~\cite{qwen3vl}: This model was fine-tuned using the same architectural configuration and data format as RynnBrain-VLA to ensure a fair comparison.

As indicated in \Cref{tab:vla_eval}, the general VLA $\pi$-0.5 struggles to identify target objects, resulting in a low RSR. This performance bottleneck stems from the limited capacity for fine-grained image-text alignment.
In contrast, while Qwen3-VL-Finetuned is derived directly from a general VLM, RynnBrain-VLA achieves superior localization accuracy and higher grasping success rates.
We attribute this advantage to our extensive pretraining on embodied pointing tasks.
Overall, RynnBrain-VLA demonstrates significantly improved success rates, notably without necessitating extensive pretraining on specific action modalities.

\begin{table*}[t]
\caption{VLA evaluation results.}
\centering
\renewcommand{\arraystretch}{1.25}
\setlength{\tabcolsep}{6.5pt}

\resizebox{\textwidth}{!}{ 
\begin{tabular}{l | ccc | ccc | ccc | ccc}
\toprule

\multirow{2}{*}{Method} & \multicolumn{3}{c|}{Pick up bread} & \multicolumn{3}{c|}{Pick up bowl} & \multicolumn{3}{c|}{Mixed} & \multicolumn{3}{c}{Overall} \\
\cline{2-13}

& PSR & RSR & SR & PSR & RSR & SR & PSR & RSR & SR & PSR & RSR & SR \\

\hline

$ \pi_{0.5} $-Finetuned & 0.7 & 0.6 & 0.5 & \textbf{0.8} & 0.5 & 0.5 & 0.5 & 0.6 & 0.4 & 0.67 & 0.57 & 0.47 \\
Qwen3-VL-Finetuned & 0.7 & \textbf{1.0} & 0.7 & 0.5 & \textbf{1.0} & 0.5 & 0.6 & \textbf{1.0} & 0.6 & 0.60 & \textbf{1.00} & 0.60 \\
\hline
\textbf{RynnBrain-VLA} & \textbf{0.8} & \textbf{1.0} & \textbf{0.8} & 0.7 & \textbf{1.0} & \textbf{0.7} & \textbf{0.9} & 0.9 & \textbf{0.8} & \textbf{0.8} & 0.97 & \textbf{0.77} \\

\bottomrule
\end{tabular}
}

\label{tab:vla_eval}
\end{table*}

\subsubsection{RynnBrain and VLA Hierarchical Evaluation}
\todo[inline, color=mybrown!20]{Assigned to: Yunxuan Mao}

\begin{figure}[!htbp]
\centering
\includegraphics[width=\textwidth]{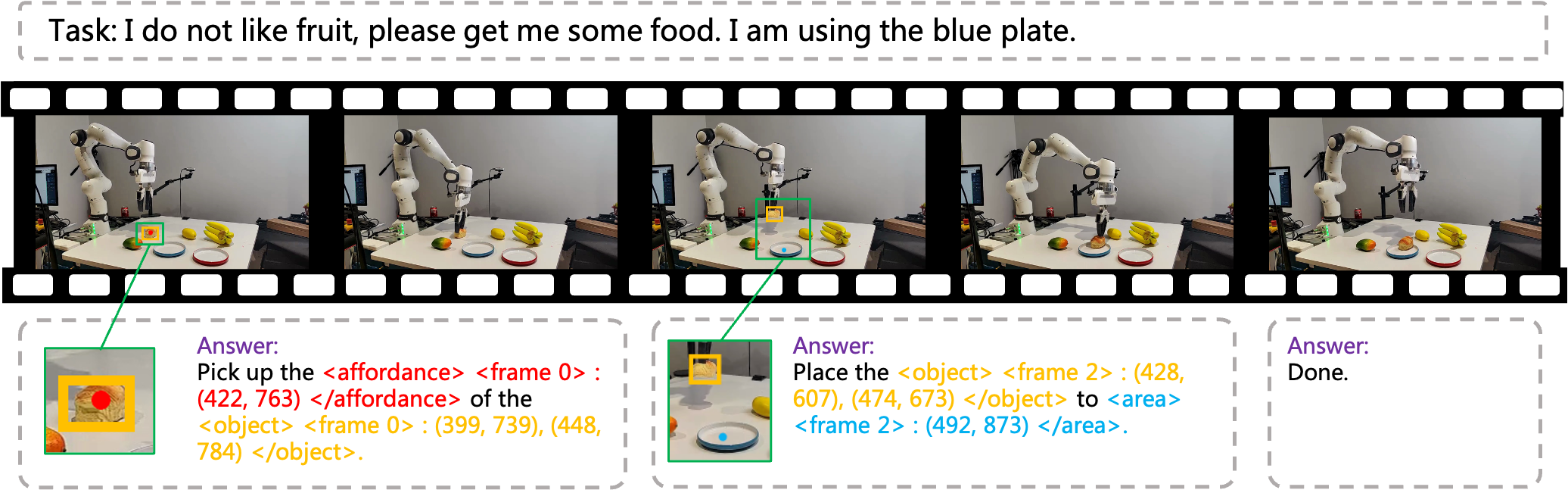}
\caption{\textbf{Planning Manipulation Video Examples of RynnBrain-Plan.} An example of the RynnBrain-Plan model on a multi-step online planning task. The executor is RynnBrain-VLA.}
\label{fig:appendix_planning_4}
\end{figure}

To validate the long-horizon planning and manipulation capabilities of our proposed framework, we integrate RynnBrain-Plan with RynnBrain-VLA to construct a hierarchical manipulation system. In this architecture, RynnBrain-Plan functions as the high-level planner, decomposing complex instructions into executable subtasks, while RynnBrain-VLA acts as the low-level controller, generating precise robot action commands. As illustrated in the qualitative results provided in the~\Cref{fig:appendix_planning_4}, these two modules are effectively integrated, demonstrating robust performance in completing long-horizon manipulation tasks.
We also evaluate a comparative setup in which Gemini generates purely textual plans that are subsequently executed by $\pi_{0.5}$. This paradigm often leads to grasping and placement mismatches in tasks involving multiple identical objects or requiring precise placement. This clearly demonstrates the significance of our physical-aware planning mode in complex scenarios and intricate operational tasks.

\section{Conclusion and Future Works}
\noindent


In this study, we introduce \textbf{RynnBrain}, a suite of advanced embodied foundation models. RynnBrain expands the capability frontier of embodied foundation models along four axes: egocentric cognition, spatio-temporal localization, physically grounded reasoning, and physics-aware planning. Across a comprehensive evaluation on 28 benchmarks, RynnBrain consistently emerges---at all model scales---as a highly capable and well-rounded open embodied foundation model. Building on the RynnBrain foundation models, we further post-train four specialized variants---RynnBrain-CoP, RynnBrain-Nav, RynnBrain-Plan, and RynnBrain-VLA---each achieving state-of-the-art performance in its respective domain and collectively demonstrating the substantial value of RynnBrain pretraining for a wide range of embodied tasks.
Beyond model development, we introduce \textbf{RynnBrain-Bench}, a high-dimensional evaluation suite designed to rigorously assess fine-grained spatio-temporal cognition and localization in embodied settings. RynnBrain-Bench advances existing benchmarks by emphasizing video understanding across episodes, precise spatio-temporal grounding, and physically meaningful pointing behaviors, providing a more faithful measure of embodied reasoning capabilities in real-world environments.

Looking forward, we view RynnBrain as a key engine for advancing multimodal foundation models into the physical world. Future embodied intelligence systems will likely comprise a holistic agent stack, including components such as a brain, cerebellum, memory modules, and a sensorimotor interface. RynnBrain is positioned to serve as a core foundation of this agent system, enabling efficient exploration, autonomous decision-making, and dynamic interaction in complex physical environments. By openly releasing the full model family under the Apache 2.0 license, we hope to empower the community to address broader embodied scenarios with RynnBrain and accelerate progress toward general embodied intelligence.

\bibliographystyle{assets/plainnat}
\bibliography{paper}

\newpage
\beginappendix


\phantomsection
\addcontentsline{toc}{section}{A. Appendix} 

\addtocontents{toc}{\protect\setcounter{tocdepth}{0}}


\appendix
\renewcommand{\thesection}{\Alph{section}}
\renewcommand{\thesubsection}{\Alph{subsection}}

\section*{Appendix Contents}
\noindent\rule{\linewidth}{0.4pt} 
\vspace{-0.5em}
\begin{itemize}[label={}, leftmargin=0.5em, itemsep=3pt]
    \item \textbf{\ref{sec:contrib} Detailed Contributions} \dotfill \pageref{sec:contrib}
    
    \item \textbf{\ref{sec:qual} Qualitative Examples} \dotfill \pageref{sec:qual}
    \begin{itemize}[label={}, leftmargin=1.5em, itemsep=1pt, topsep=0pt]
        \item \small \ref{sec:qual_cog} Examples for Embodied Cognition \dotfill \pageref{sec:qual_cog}
        \item \small \ref{sec:qual_loc} Examples for Embodied Location \dotfill \pageref{sec:qual_loc}
        \item \small \ref{sec:qual_vis} Examples for General Visual Understanding \dotfill \pageref{sec:qual_vis}
        \item \small \ref{sec:qual_phy} Examples for Physically Grounded Reasoning \dotfill \pageref{sec:qual_phy}
        \item \small \ref{sec:qual_nav} Examples for Navigation \dotfill \pageref{sec:qual_nav}
        \item \small \ref{sec:qual_plan} Examples for Manipulation Planning \dotfill \pageref{sec:qual_plan}
    \end{itemize}

    \item \textbf{\ref{sec:prompts} Prompts Details} \dotfill \pageref{sec:prompts}
    \begin{itemize}[label={}, leftmargin=1.5em, itemsep=1pt, topsep=0pt]
        \item \small \ref{sec:prompt_train} Training QA Prompts \dotfill \pageref{sec:prompt_train}
        \item \small \ref{sec:prompt_eval} Evaluation and Inference Prompts \dotfill \pageref{sec:prompt_eval}
        \item \small \ref{sec:hyper} Hyper-parameters for Evaluation \dotfill \pageref{sec:hyper}
    \end{itemize}
\end{itemize}
\vspace{-0.5em}
\noindent\rule{\linewidth}{0.4pt} 
\vspace{1em}

\subsection{Detailed Contributions}
\label{sec:contrib} 
\begin{itemize}
    \item \textbf{Data}: Ronghao Dang, Zhikai Wang, Yunxuan Mao, Yuqian Yuan,  Bohan Hou, Jiangpin Liu, Kehan Li, Jiayan Guo, Xin Li, Sicong Leng, Minghao Zhu, Yang Bai, Qian Jiang
    \item \textbf{Model Training}:
    \begin{itemize}[label=$\blacksquare$]
        \item \textbf{Pre-training}: Jiayan Guo, Kehan Li, Ronghao Dang, Sicong Leng, Zhikai Wang, Xin Li, Jiangpin Liu, Yuqian Yuan, Yunxuan Mao
        \item \textbf{Post-Training}: 
        \begin{itemize}[label=--]
            \item \textbf{Chain-of-Point Reasoning}: Jiayan Guo, Zhikai Wang, Minghao Zhu, Yuqian Yuan, Ronghao Dang, Kehan Li
            \item \textbf{Vision-Language Navigation}: Jiangpin Liu, Yuqian Yuan, Kehan Li, Ronghao Dang, Liuyi Wang
            \item \textbf{Planning}: Yunxuan Mao, Bohan Hou, Kehan Li, Ronghao Dang, Yuqian Yuan, Xiao Lin
            \item \textbf{VLA}: Kehan Li, Bohan Hou, Yunxuan Mao, Yuqian Yuan, Jiayan Guo, Ronghao Dang
        \end{itemize}
    \end{itemize}
    \item \textbf{Infrastructure}: Kehan Li, Jiayan Guo, Xin Li
    \item \textbf{Evaluation}: Minghao Zhu, Bohan Hou, Yuqian Yuan, Sicong Leng, Jiayan Guo, Kehan Li, Yunxuan Mao, Jiangpin Liu, Ronghao Dang, Yuming Jiang, Xin Li
    \item \textbf{Hardware and Robot System}: Yaxi Zhao, Minghua Zeng, Junlong Gao
    \item \textbf{Senior Advisory}: Jun Cen, Siteng Huang, Wenqiao Zhang (Zhejiang University), Chengju Liu (Tongji University), Jianfei Yang (Nanyang Technological University), Shijian Lu (Nanyang Technological University), Deli Zhao
\end{itemize}

\subsection{Qualitative Examples}
\label{sec:qual}
This section presents an extensive set of visual examples to demonstrate RynnBrain’s robust capabilities across a wide range of embodied tasks.

\subsubsection{Examples for Embodied Cognition}
\label{sec:qual_cog}
\todo[inline, color=myred!20]{Assigned to: Yuqian Yuan}

\begin{figure}[!htbp]
\centering
\includegraphics[width=\textwidth]{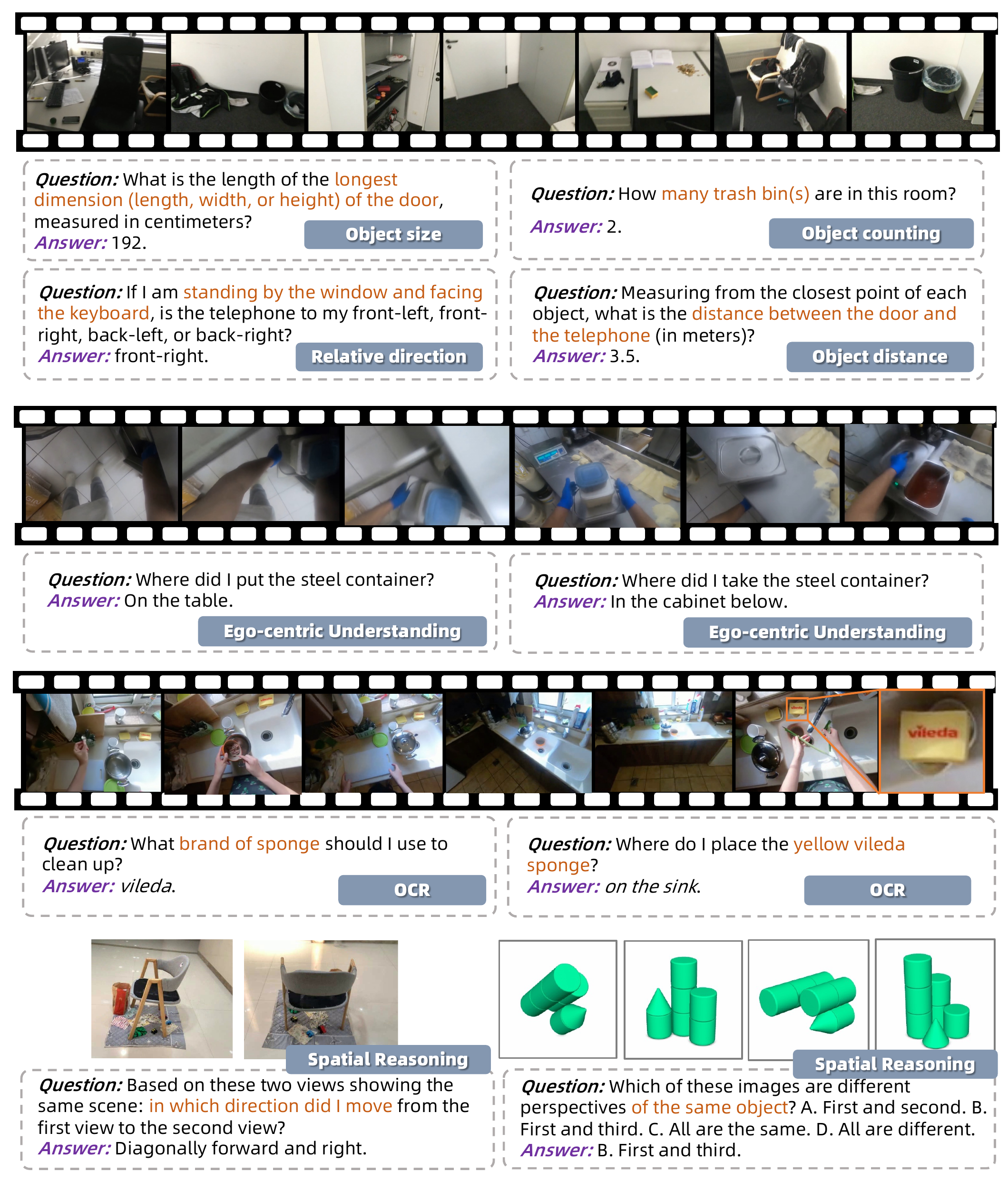}
\caption{\textbf{Embodied Cognition Examples of RynnBrain.} RynnBrain supports diverse embodied cognition tasks, including spatial understanding of object size, direction, distance, and counting, OCR-based perception, and higher-level spatial reasoning over egocentric views and 3D shapes.} 
\label{fig:example1}
\end{figure}

\begin{figure}[!htbp]
\centering
\includegraphics[width=\textwidth]{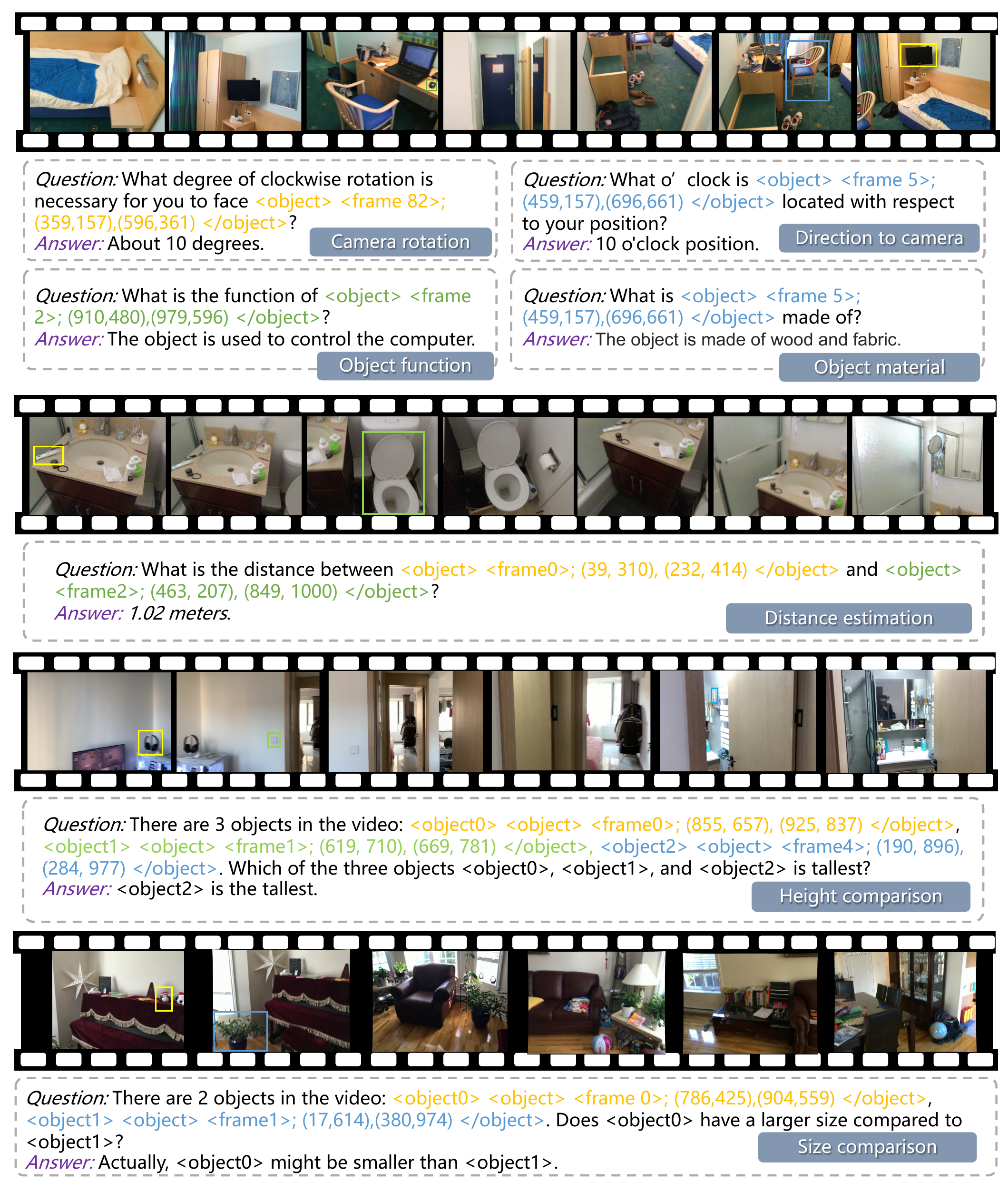}
\caption{\textbf{Embodied Cognition Examples of RynnBrain.} RynnBrain also supports a diverse range of fine-grained embodied cognition tasks, including spatial understanding of camera rotation, direction, distance, and size, as well as object understanding of function and material.} 
\label{fig:example1_2}
\end{figure}

As shown in \Cref{fig:example1} and \Cref{fig:example1_2}, our RynnBrain supports a wide range of embodied cognition abilities, including estimating object size and distance, reasoning about relative directions and object counts, performing fine-grained grounded object-centric understanding and OCR-based perception, and conducting higher-level spatial reasoning over egocentric views and 3D shapes.

\subsubsection{Examples for Embodied Location}
\label{sec:qual_loc}
\todo[inline, color=mygreen!20]{Assigned to: Zhikai Wang}

\begin{figure}[!htbp]
\centering
\includegraphics[width=\textwidth]{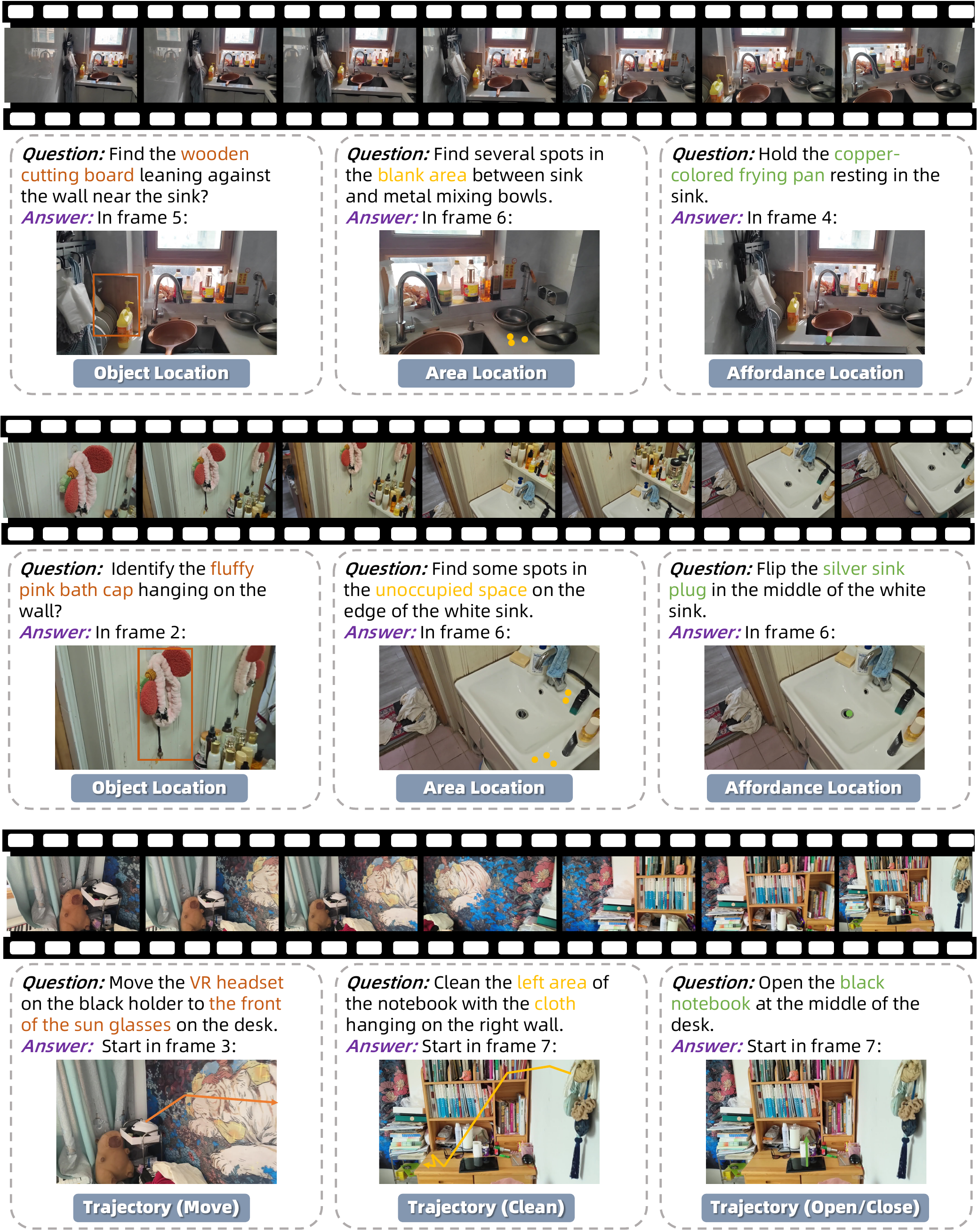}
\caption{\textbf{Embodied Location Video Examples of RynnBrain.} RynnBrain excels at grounded spatial reasoning, supporting video-based location tasks for object, area, affordance, trajectory. These examples highlight its ability to map linguistic descriptions to 3D locations and actions in real-world scenes.}
\label{fig:example2}
\end{figure}

\begin{figure}[!htbp]
\centering
\includegraphics[width=\textwidth]{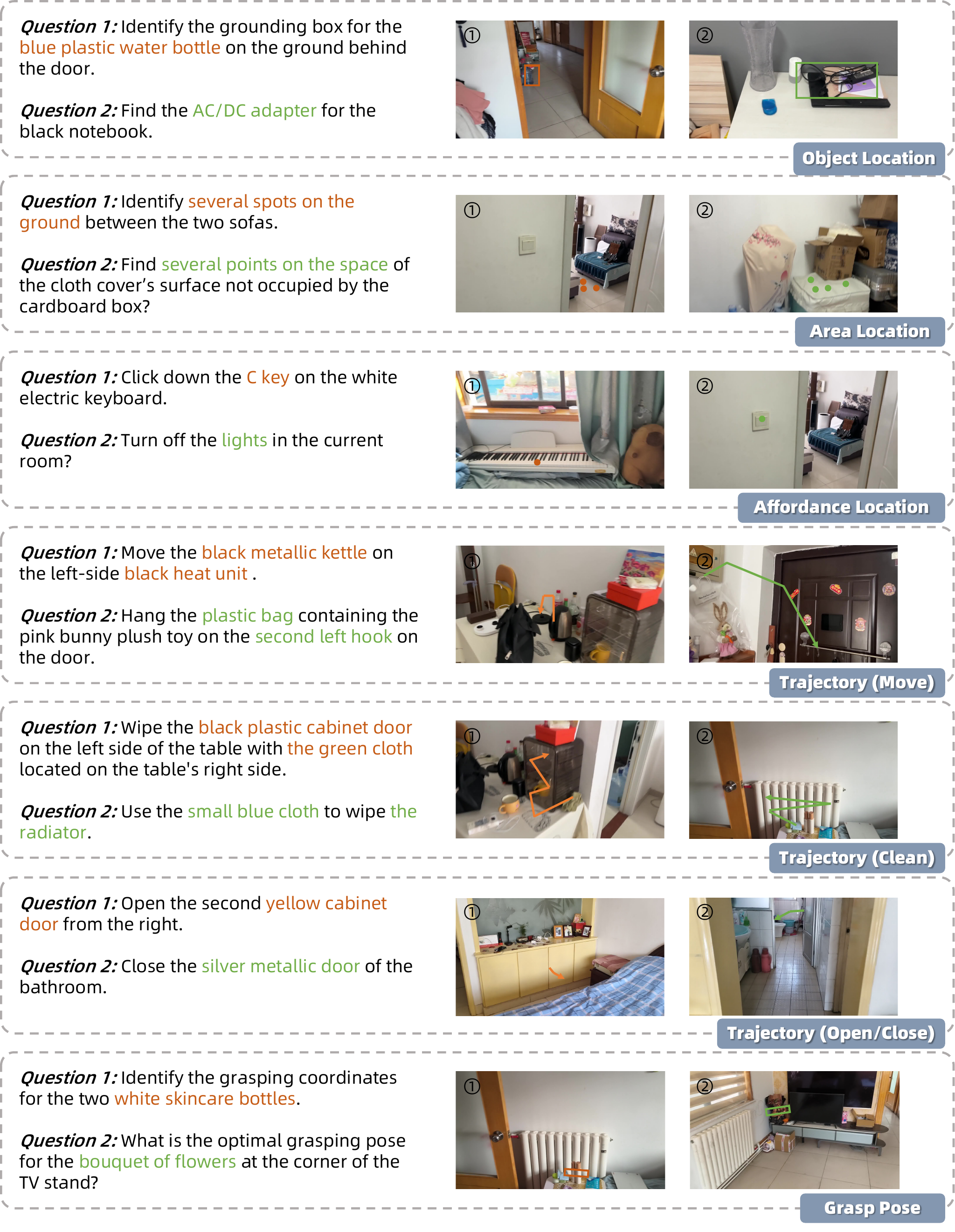}
\caption{\textbf{Embodied Location Image Examples of RynnBrain.} RynnBrain excels at grounded spatial reasoning, supporting image-based location tasks for object, area, affordance, trajectory, and grasp pose. These examples highlight its ability to map linguistic descriptions to 3D locations and actions in real-world scenes.}
\label{fig:example2_2}
\end{figure}

As illustrated in \Cref{fig:example2} and \Cref{fig:example2_2}, our RynnBrain demonstrates robust embodied location understanding by accurately interpreting spatial references in egocentric views and grounding natural language instructions to precise physical coordinates. It can localize objects based on relative positions, identify graspable items via functional cues, and generate complex spatial trajectories such as moving or cleaning specified regions.

\subsubsection{Examples for General Visual Understanding}
\label{sec:qual_vis}
\todo[inline, color=myred!20]{Assigned to: Yuqian Yuan}

\begin{figure}[!htbp]
\centering
\includegraphics[width=\textwidth]{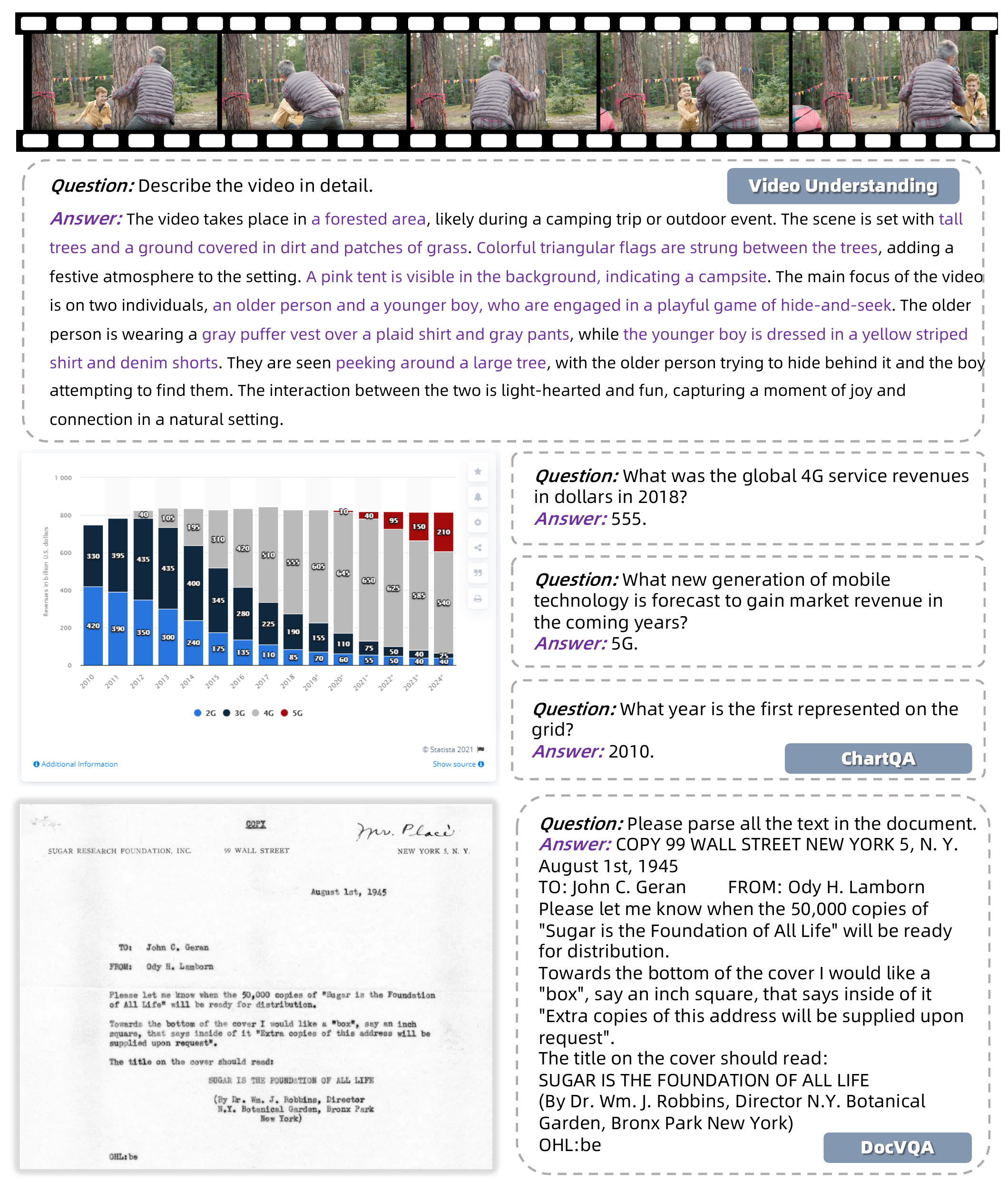}
\caption{\textbf{General Visual Understanding Examples of RynnBrain.} RynnBrain also preserves strong general visual understanding capabilities, including video comprehension and image understanding for content such as charts and documents.}
\label{fig:example3}
\end{figure}

As illustrated in \Cref{fig:example3}, RynnBrain also demonstrates strong general visual understanding abilities, including video comprehension and image-level understanding of structured content such as charts and documents.

\subsubsection{Examples for physically Grounded Reasoning}
\label{sec:qual_phy}
\begin{figure}[!htbp]
\centering
\includegraphics[width=\textwidth]{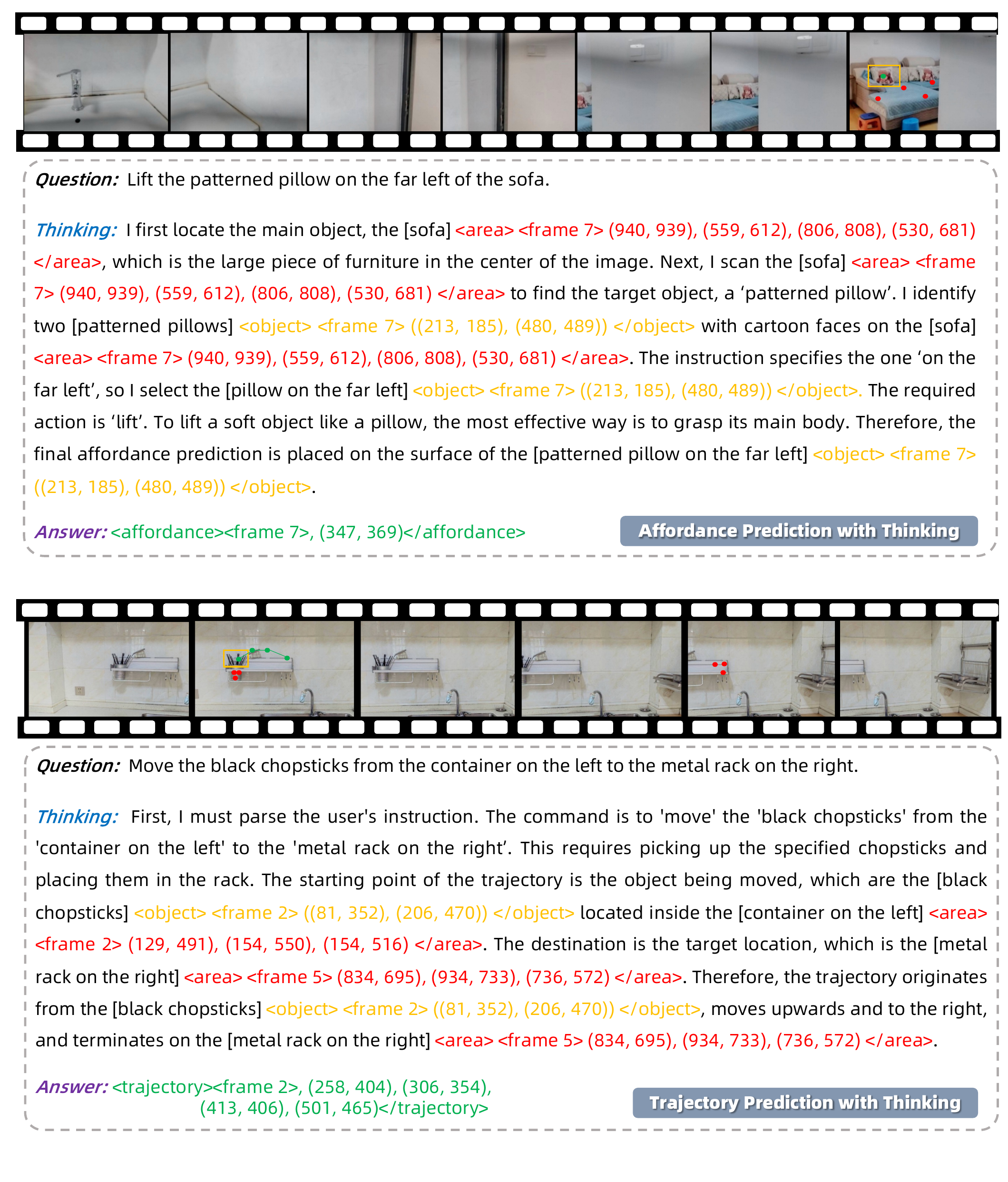}
\caption{\textbf{Embodied Location Video Examples of RynnBrain.} RynnBrain excels at grounded spatial reasoning, supporting video-based location tasks for object, area, affordance, trajectory. These examples highlight its ability to map linguistic descriptions to 3D locations and actions in real-world scenes.}
\label{fig:thinking_example}
\end{figure}

\todo[inline, color=myblue]{Assigned to: Jiayan Guo}
As illustrated in~\Cref{fig:thinking_example}, RynnBrain-CoP demonstrates physically grounded reasoning capabilities.

\subsubsection{Examples for Navigation}
\begin{figure}[!htbp]
\centering
\includegraphics[width=\textwidth]{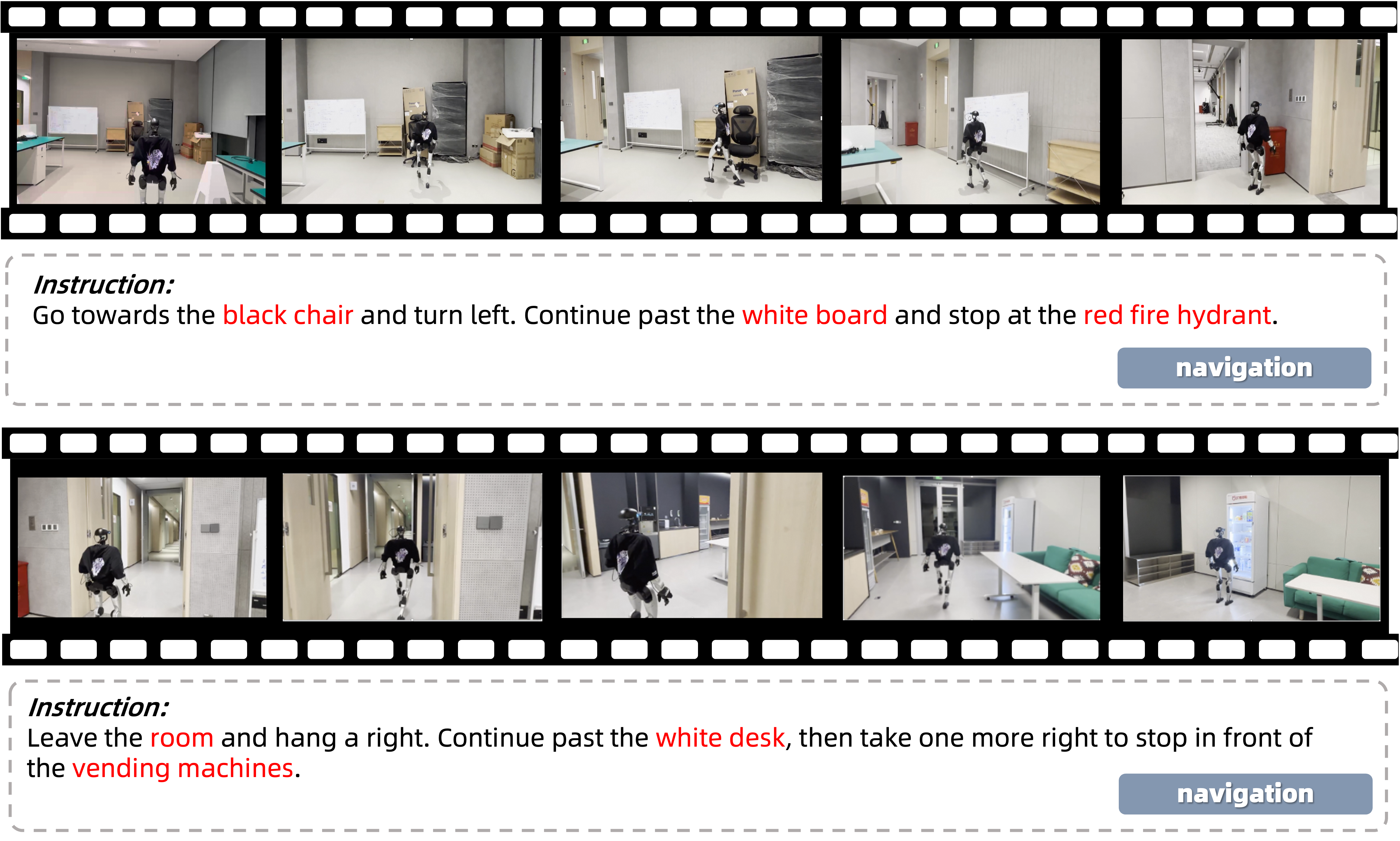}
\caption{\textbf{Visual Language Navigation Video Examples of RynnBrain in Real Environment.} Examples of the RynnBrain-Nav model in a real indoor environment. The results demonstrate the strong navigation ability of the  model in the real environment.}
\label{fig:real_example}
\end{figure}

\begin{figure}[!htbp]
\centering
\includegraphics[width=\textwidth]{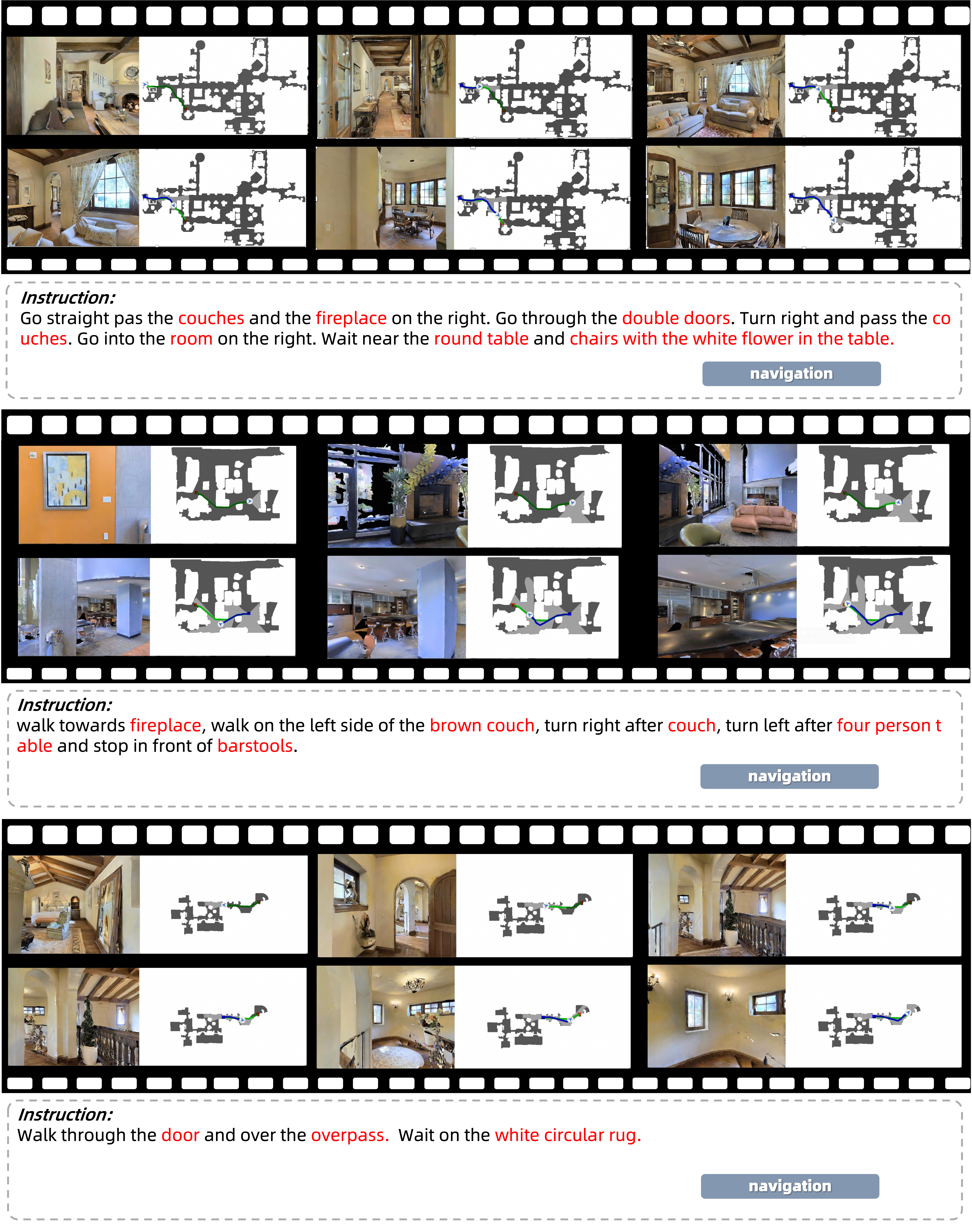}
\caption{\textbf{Visual Language Navigation Video Examples of RynnBrain on R2R-CE.} Examples of the RynnBrain-Nav model on R2R-CE. The results demonstrate the strong navigation ability of the  model.}
\label{fig:r2r_example}
\end{figure}

\begin{figure}[!htbp]
\centering
\includegraphics[width=\textwidth]{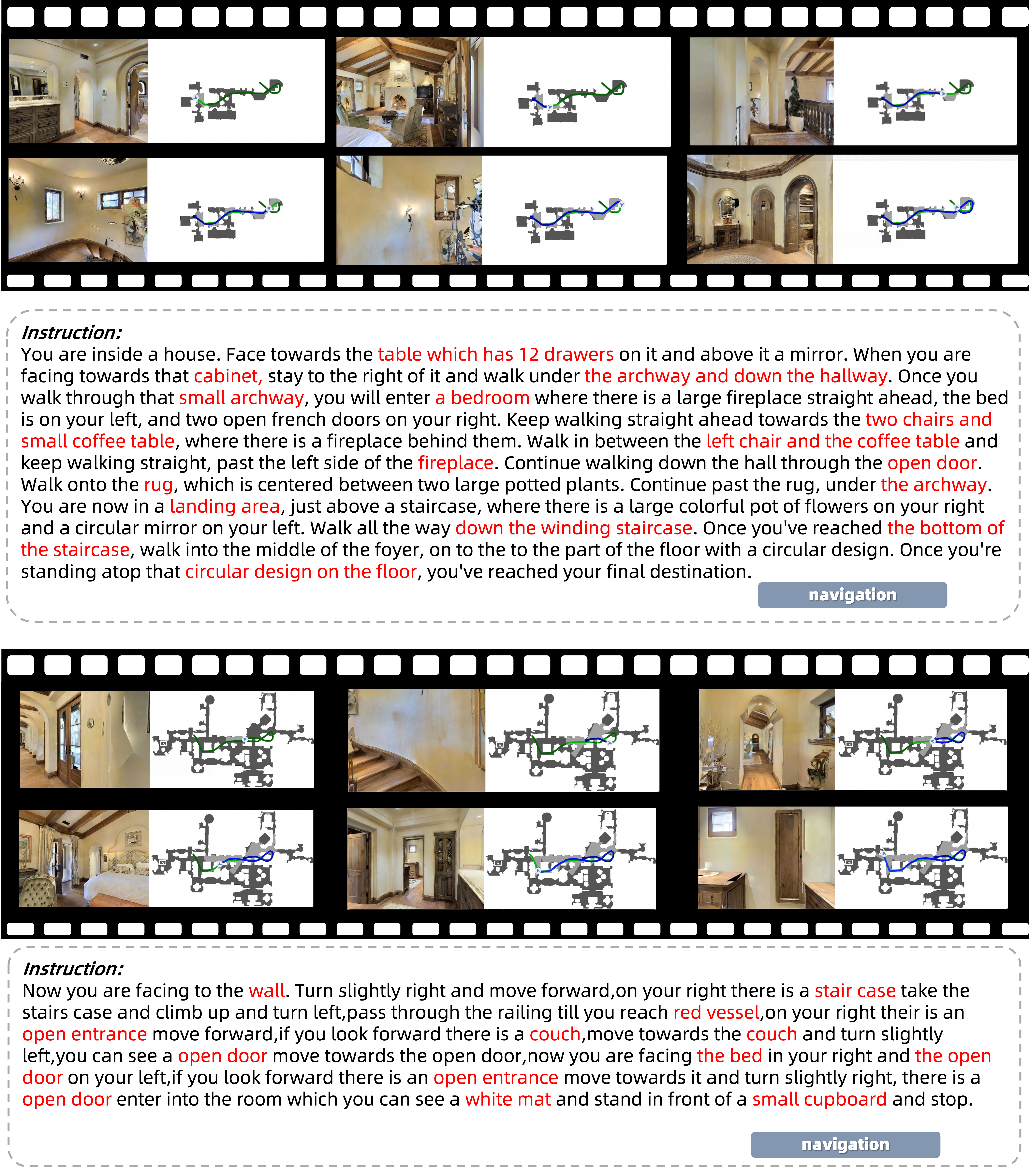}
\caption{\textbf{Visual Language Navigation Video Examples of RynnBrain on RxR-CE.} Examples of the RynnBrain-Nav model on RxR-CE. The results demonstrate the strong navigation ability of the  model.}
\label{fig:rxr_example}
\end{figure}

\todo[inline, color=mypurple!20]{Assigned to: Jiangpin Liu}
\label{sec:qual_nav}
As illustrated in~\Cref{fig:real_example},~\Cref{fig:r2r_example} and~\Cref{fig:rxr_example}, RynnBrain-Plan demonstrates robust long-horizon navigation planning capabilities. 

\subsubsection{Examples for Manipulation Planning}
\begin{figure}[!htbp]
\centering
\includegraphics[width=0.9\textwidth]{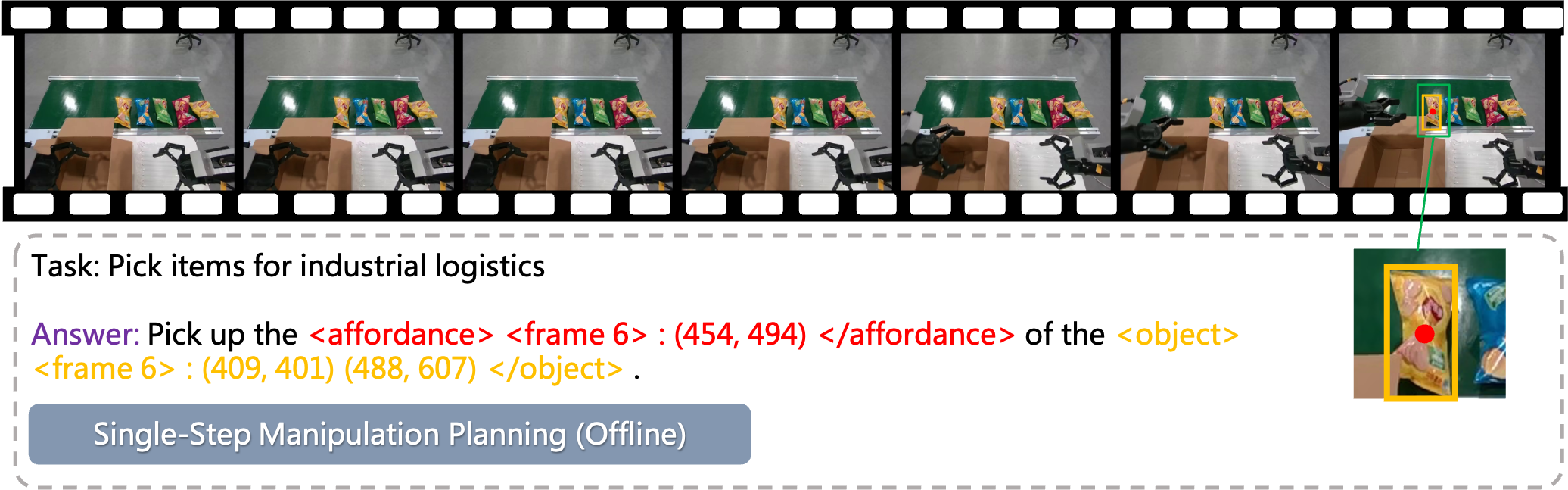}
\caption{\textbf{Planning Manipulation Video Examples of RynnBrain-Plan.} An example of the RynnBrain-Plan model on a one-step offline planning task on the Agibot Dataset.}
\label{fig:appendix_planning_1}
\end{figure}

\begin{figure}[!htbp]
\centering
\includegraphics[width=0.9\textwidth]{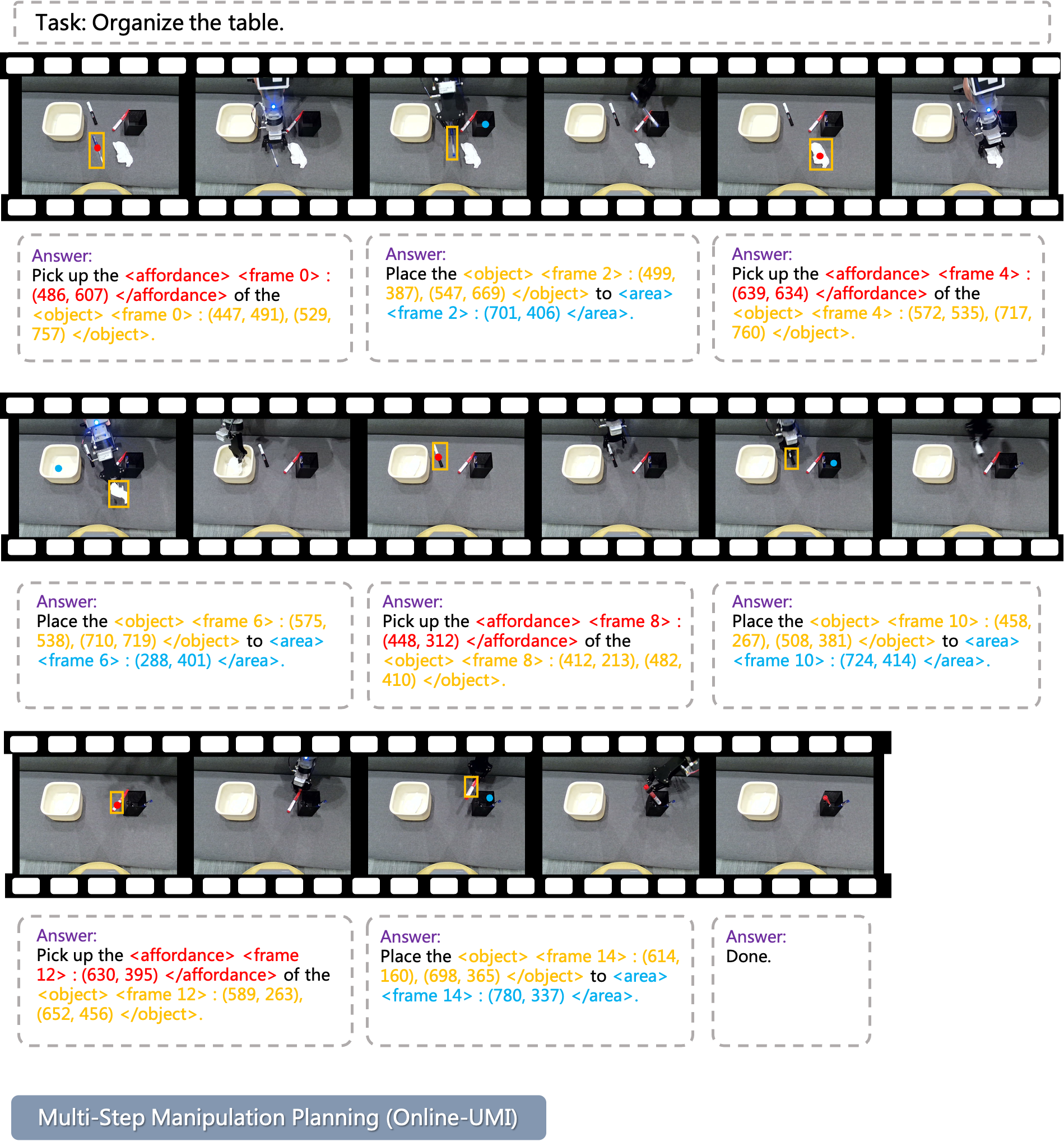}
\caption{\textbf{Planning Manipulation Video Examples of RynnBrain-Plan.} An example of the RynnBrain-Plan model on a multi-step online planning task. The executer is a human expert with a UMI.}
\label{fig:appendix_planning_2}
\end{figure}

\begin{figure}[!htbp]
\centering
\includegraphics[width=0.9\textwidth]{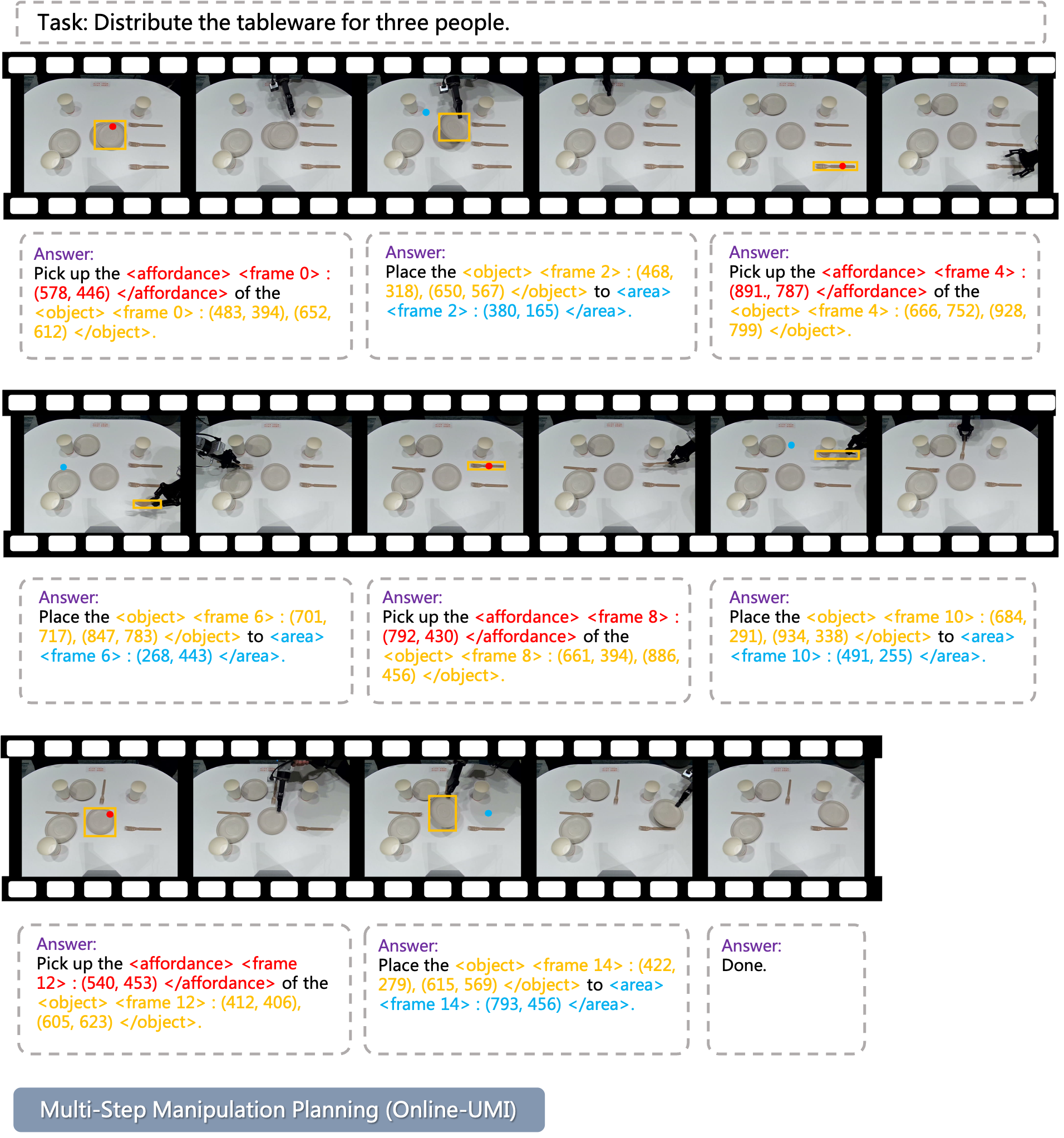}
\caption{\textbf{Planning Manipulation Video Examples of RynnBrain-Plan.} An example of the RynnBrain-Plan model on a multi-step online planning task. The executer is a human expert with a UMI.}
\label{fig:appendix_planning_3}
\end{figure}

\todo[inline, color=mybrown!20]{Assigned to: Yunxuan Mao}
\label{sec:qual_plan}
As illustrated in~\Cref{fig:appendix_planning_1},~\Cref{fig:appendix_planning_2} and~\Cref{fig:appendix_planning_3}, RynnBrain-Plan demonstrates robust long-horizon planning capabilities. Furthermore, its precise grounding ability enables our method to handle a wide range of fine-grained manipulation tasks adeptly. The details of each task are as follows:

\paragraph{Distribute Tableware.} In this task, the planning model is required to distribute tableware for a specified number of people. The detailed task prompt is provided below.
\begin{itemize}
    \item Easy: Distribute the tableware on the table among three people.
    \item Medium: Distribute the tableware on the table among three people, making sure that the cups are on the right side of each person.
    \item Hard: Distribute the tableware on the table among four people, making sure that the cups are on the right side of each person.
    
\end{itemize}

\paragraph{Object Classification.} In this task, the planning model is required to categorize the food items on the table and arrange them in a row according to specific instructions. The detailed task prompt is provided below.
\begin{itemize}
    \item Easy: Sort the fruits on the table and arrange each type of fruit in a row.
    \item Medium:  Sort the fruits on the table into categories and arrange each category in a row, placing them on either side of the mug.
    \item Hard: Sort the fruits on the table into categories and arrange each category in a row, placing them on either side of the mug. Don't move the fruits in the bowl.
    
\end{itemize}

\paragraph{Desk Organization.} In this task, the planning model is required to place different types of pens and garbage into designated locations according to specific requirements. The detailed task prompt is provided below.

\begin{itemize}
    \item Easy: Tidy up the desktop. Put the thin pens in the pen holder and arrange the thick pens from left to right in the order of red, and black. Finally, make sure there is no trash on the desktop.
    \item Medium:  Tidy up the desktop. Put the thin pens in the pen holder and arrange the thick pens from left to right in the order of red, black and blue. Finally, make sure there is no trash or used paper cups on the table.
    \item Hard: Tidy up the desktop. Put the thin pens in the pen holder and arrange the thick pens from left to right in the order of red, black and blue. Finally, make sure there is no trash or used paper cups on the table.
    
\end{itemize}

\paragraph{Table Bussing (OOD).} In this task, the planning model is required to generate a detailed action plan in response to the instruction ``Bus the table''. The detailed task setting is provided below.

\begin{itemize}
    \item Easy: Two forks, two pens, one plate, and one pen holder. 
    \item Medium: Two forks, two pens, two plates, two cups and one pen holder. 
    \item Hard: Two forks, two pens, two plates, two cups, one trash, one trash can and one pen holder. 
    
\end{itemize}

\clearpage

\subsection{Prompts Details}
\label{sec:prompts}
\subsubsection{Training QA Prompts}
\label{sec:prompt_train}
To ensure reproducibility, we list the detailed prompt templates used for each training task in \Cref{tab:train_prompts}. All prompts are presented in a unified Python f-string format.

\subsubsection{Evaluation and Inference Prompts}
\label{sec:prompt_eval}
To ensure reproducibility, we list the detailed prompt templates used for each benchmark in \Cref{tab:prompts}. All prompts are presented in a unified Python f-string format.

{ 
\small
\renewcommand{\arraystretch}{1.3}

\begin{xltabular}{\linewidth}{@{}l X@{}}
    \caption{Prompt Templates for Various Training Tasks} \label{tab:train_prompts} \\
    \toprule
    \textbf{Training Task} & \textbf{Prompt Template (Unified Format)} \\
    \midrule
    \endfirsthead

    \multicolumn{2}{c}{{\bfseries \tablename\ \thetable{} -- continued from previous page}} \\
    \toprule
    \textbf{Training Task} & \textbf{Prompt Template (Continued)} \\
    \midrule
    \endhead

    \midrule
    \multicolumn{2}{r}{{Continued on next page...}} \\
    \endfoot

    \bottomrule
    \endlastfoot


    \makecell[l]{Object Understanding} 
    &  \texttt{f"I'd like to know about the area labeled <object> (\{x[0]\}, \{y[0]\}), (\{x[1]\}, \{y[1]\}) </object> in the image. Can you give a short description?"}
       \\
\midrule
      \makecell[l]{Spatial Understanding} 
    & \texttt{f"You are in the last frame of the video. There are \{n\} objects in the video: <object\{idx\}> <object>  <frame \{frame\_id\}>: (\{x[idx][0]\}, \{y[idx][0]\}), (\{x[idx][1]\}, \{y[idx][1]\}) </object>, ... \textbackslash n\{question\}"} \\
\midrule
    \makecell[l]{OCR} 
    & \texttt{f"Locate the text "\{text\}" in this video.\textbackslash n\textbackslash nStep 1: Predict the key frame.\textbackslash nStep 2: Output a tuple series.\textbackslash nOutput format: <area> <frame n>: (x1, y1), (x2, y2), .... </area>\textbackslash n with all coordinates normalized to 0-1000 range."}\\
\midrule
    \makecell[l]{Object Location} 
    & 1. \texttt{f"\{question\}\textbackslash nOutput the bounding box in the format <object> <frame n>: ...; (x1,y1), (x2,y2) </object>. n is the chosen frame index."}\newline
    2. \texttt{f"\{question\}\textbackslash nGenerate coordinates for one object bounding box. Constraints: x1,y1,x2,y2 $\in$ [0,1000]. Response must be in the format: <object> (x1, y1), (x2, y2) </object>"}\\
\midrule
    \makecell[l]{Area Location} 
    & 1. \texttt{f"\{question\}\textbackslash nFirst perform key frame prediction, then generate a sequence of coordinate tuples.\textbackslash nOutput format: <area> <frame n>: ...; (x1, y1), (x2, y2), .... </area>\textbackslash n Each coordinate pair must contain normalized pixel values within the [0, 1000] range."}\newline
    2. \texttt{f"\{question\}\textbackslash nExpress the coordinates as a tuple sequence in the format <area> (x1, y1), (x2, y2), ... </area> with all coordinate values normalized to the standardized pixel coordinate system spanning 0 to 1000."}
    \\
\midrule
    \makecell[l]{Affordance Location} 
    & 1. \texttt{f"\{question\}\textbackslash n1. First identify the key frame\textbackslash n2. Then predict one affordance point.\textbackslash nOutput format: <affordance> <frame n>: ...; (x, y) </affordance>\textbackslash nCoordinates normalized to 0-1000 pixel space."}\newline
    2. \texttt{f"\{question\}\textbackslash nTask: Affordance point prediction\textbackslash n- Identify one possible affordance point\textbackslash n- Normalize coordinates to 0-1000 range\textbackslash n- Output format: <affordance> (x, y) </affordance>\textbackslash n- Example: [450, 320]"}
    \\
\midrule
    \makecell[l]{Trajectory Location} 
    & 1. \texttt{f"\{question\}\textbackslash nor trajectory completion:\textbackslash n1. First locate the frame with the trajectory start point\textbackslash n2. Then predict up to 10 key points as list. Output format: <trajectory> <frame n>: ...; (x1, y1), (x2, y2), .... </trajectory>\textbackslash nAll coordinates normalized to 0-1000 pixel space."}\newline
    2. \texttt{f"\{question\}\textbackslash nPredict a trajectory comprising up to 10 key points. Return coordinates in the format <trajectory> (x1, y1), (x2, y2), ... </trajectory> with all values normalized to the [0, 1000] range."}
    \\
\midrule
    \makecell[l]{Navigation} 
    & \texttt{f"You are an autonomous navigation assistant. Your task is to Walk straight, and when you reach the end of the table turn right.  Wait by the brass chairs. Devise an action sequence to follow the instruction using the four actions: TURN LEFT, TURN RIGHT, MOVE FORWARD, or STOP."}
    \\
\midrule
    \makecell[l]{Planning} 
    & \texttt{f"You are a sophisticated dual-arm robot planning the next action for the goal: \{question\}. Put the thin pens in the pen holder and arrange the thick pens from left to right in the order of red, black and blue. Finally, make sure there is no trash on the desktop..\textbackslash n\textbackslash n        Adhere to the following output rules:\textbackslash n        - Rule 1: The response must be a single, complete sentence.\textbackslash n        - Rule 2: The sentence must embed data by selecting a frame `n' and predicting integer coordinates within the [0, 1000] range.\textbackslash n        - Rule 3: Data format within tags must be `<tag> <frame n>: (data) </tag>', where `data' is a single point for both affordance and area, and two points `(min\_coord), (max\_coord)' for object."}\\

    \end{xltabular}
}

{ 
\small
\renewcommand{\arraystretch}{1.3}

\begin{xltabular}{\linewidth}{@{}l X@{}}
    \caption{Prompt Templates for Various Benchmarks} \label{tab:prompts} \\
    \toprule
    \textbf{Benchmark} & \textbf{Prompt Template (Unified Format)} \\
    \midrule
    \endfirsthead

    \multicolumn{2}{c}{{\bfseries \tablename\ \thetable{} -- continued from previous page}} \\
    \toprule
    \textbf{Benchmark} & \textbf{Prompt Template (Continued)} \\
    \midrule
    \endhead

    \midrule
    \multicolumn{2}{r}{{Continued on next page...}} \\
    \endfoot

    \bottomrule
    \endlastfoot


    \makecell[l]{VSI-Bench} 
    & 1. \texttt{f"\{question\}\textbackslash nAnswer with the option's letter from the given choices directly."} \newline
      2. \texttt{f"\{question\}\textbackslash n\textbackslash nAnswer the question with an exact number, which should be accurate to at most two decimal places."} \\
    \midrule
    
    MMSI & \texttt{f"\{question\}"} \\
    \midrule
    
    ERQA & \texttt{f"\{question\}\textbackslash nAnswer with the option letter from the given choices directly."} \\
    \midrule
    
    \makecell[l]{RoboSpatial} 
    & 1. \texttt{f"\{question\}. Pinpoint several points within the vacant space situated to the left of the vacuum. Your answer should be formatted as a list of tuples, i.e. [(x1, y1), ...], where each tuple contains the x and y coordinates... indicating the normalized pixel locations of the points."} \newline
      2. \texttt{f"\{question\} Answer yes or no."} \\
    \midrule
    
    EgoTaskQA & \texttt{f"Select the best answer to the following multiple-choice question based on the video.\textbackslash n\{question\}\textbackslash nOptions:\textbackslash n(A) \{options[0]\}\textbackslash n(B) \{options[1]\}\textbackslash n(C) \{options[2]\}\textbackslash n(D) \{options[3]\}\textbackslash n(E) \{options[4]\}\textbackslash nAnswer with the option's letter from the given choices directly and only give the best option. The best answer is: "} \\
    \midrule
    
    EgoTextVQA\_indoor & \texttt{f"You are a person in the situation shown in the following consecutive images... Answer the question as detailed as possible, covering all relevant aspects and providing comprehensive context.\textbackslash n\textbackslash nQuestion: \{question\}"} \\
    \midrule
    
    Open-X VQA & \texttt{f"Select the best answer to the following multiple-choice question based on the image.\textbackslash n\{question\}\textbackslash nOptions:\textbackslash n\{opts\_text\}\textbackslash nAnswer with the option's letter from the given choices directly and only give the best option. The best answer is: "} \\
    \midrule
    
    QAEgo4D & \texttt{f"You are a helpful assistant. Please evaluate the predicted answer based on the given question. A score of 0 means the answer is completely incorrect... Output only a single score from the following set: [0, 1, 2, 3, 4, 5]. \{question\}"} \\
    \midrule
    
    MindCube & \texttt{f"Select the best answer to the following multiple-choice question based on the image.\textbackslash n\{question\}\textbackslash nAnswer with the option's letter from the given choices directly and only give the best option. The best answer is: "} \\
    \midrule
    
    RynnBrain-Object & \texttt{f"\{question\} Your current position is at the last frame of the video."} \\
    \midrule

    RynnBrain-Spatial & \texttt{f"\{question\}"} \\
    \midrule

    RefSpatial-Bench & \texttt{f"Locate \{object\_name\} in this image. Output the point coordinates in JSON format."} \\
    \midrule
    
    ShareRobot-Affordance & \texttt{f"\{question\}The coordinates should be between 0 and 1000, indicating the normalized pixel locations of the point."} \\
    \midrule
    
    ShareRobot-Trajectory & \texttt{f"\{question\}Your answer should be formatted as a list of tuples, i.e. [(x1, y1), (x2, y2), ...], where each tuple contains the x and y coordinates of a point. The coordinates should be between 0 and 1000, indicating the normalized pixel locations of the point."} \\
    \midrule
    
    \makecell[l]{Cornell-Grasp \\ VMRD-Grasp} & \texttt{f"How should the robot grasp the object? Output the grasping pose as 4 corner points of the gripper rectangle.\textbackslash n- Format: <grasp pose> (x1, y1), (x2, y2), (x3, y3), (x4, y4) </grasp pose>\textbackslash n- All coordinates in range [0, 1000] (normalized)\textbackslash n- The 4 corners define the gripper's position, orientation, and width"} \\
    \midrule
    
    RynnBrain-Grounding & \texttt{f"\{question\}. Output the bounding box in the format <object> <frame n>: ...; (x1,y1), (x2,y2) </object>. n is the chosen frame index."} \\
    \midrule

    RynnBrain-Area & \texttt{f"\{question\}. First predict the key frame, then output coordinates as a series of tuples. \textbackslash nOutput format: <area> <frame n>: ...; (x1, y1), (x2, y2), .... </area>\textbackslash n All coordinates must be normalized between 0 and 1000."} \\
    \midrule

    RynnBrain-Affordance & \texttt{f"\{question\}. First predict the key frame, then output a single affordance point as coordinates (x, y).\textbackslash nOutput format: <affordance> <frame n>: ...; (x, y) </affordance>\textbackslash n Both x and y values must be normalized between 0 and 1000."} \\
    \midrule

    RynnBrain-Trajectory & \texttt{f"\{question\}. First predict the frame containing the trajectory start point, then output up to 10 key trajectory points as a list of tuples in the format: <trajectory> <frame n>: ...; (x1, y1), (x2, y2), .... </trajectory> All coordinates must be normalized between 0 and 1000."} \\
    \midrule

    AI2D & \texttt{f"\{question\}"} \\
    \midrule
    
    ChartQA & \texttt{f"\{question\}\textbackslash nAnswer the question using a single word or phrase."} \\
    \midrule
    
    DocVQA & \texttt{f"\{question\}\textbackslash nAnswer the question with a single word or phrase."} \\
    \midrule
    
    MVBench & \texttt{f"Question: \{question\}\textbackslash nOptions:\textbackslash n\{option\_string\}Answer with the option's letter from the given choices directly and only give the best option."} \\
    \midrule
    
    RealworldQA & \texttt{f"\{question\}"} \\
    \midrule
    
    InfoVQA\_test & \texttt{f"\{question\}\textbackslash nAnswer the question with a single word or phrase."} \\
    \midrule
    
    EgoSchema & \texttt{f"Select the best answer to the following multiple-choice question based on the video.\textbackslash n\{question\}\textbackslash nOptions:\textbackslash n(A) \{options[0]\}\textbackslash n(B) \{options[1]\}\textbackslash n(C) \{options[2]\}\textbackslash n(D) \{options[3]\}\textbackslash n(E) \{options[4]\}\textbackslash nAnswer with the option's letter from the given choices directly and only give the best option. The best answer is: "} \\
    \midrule
    
    VideoMME w/o sub & \texttt{f"Select the best answer to the following multiple-choice question based on the video. Respond with only the letter (A, B, C, or D) of the correct option.\textbackslash n \{question\} "} \\

\end{xltabular}
}

\subsubsection{Hyper-parameters for Evaluation}
\label{sec:hyper}
\todo[inline, color=mygray!20]{Assigned to: Kehan Li}

To ensure reproducibility, we disabled sampling during autoregressive text generation unless otherwise specified.
For images processing, we constrained the resolution by setting min\_pixels to $ 16 \times 32 \times 32 $ and max\_pixels to $ 16384 \times 32 \times 32 $.
For video-related benchmarks, frames are sampled at 2 FPS; if the total exceeds 512 frames, we apply uniform sampling to maintain a maximum of 512.
The min\_pixels for each frame and max\_pixels for the whole video are set to $ 16 \times 32 \times 32 $ and $ 24576 \times 32 \times 32 $, respectively.
On certain pointing-related benchmarks (e.g., ERQA, RoboSpatial, and ShareRobot), we observed that sampling can further enhance model performance.
In these instances, we set the temperature to 0.2, top\_p to 0.95, and top\_k to 50.
Additionally, for RefSpatial and ShareRobot—which involve numerous precise positioning tasks—we employed a higher resolution by increasing min\_pixels to $ 1024 \times 32 \times 32 $.

\end{document}